\documentclass{article}
\usepackage{microtype}
\usepackage{graphicx}
\usepackage{subfigure}
\usepackage{booktabs} 
\usepackage{float}
\usepackage{caption}
\usepackage{enumitem}
\usepackage{longtable}
\usepackage{array} 
\usepackage{dsfont}
\usepackage{hyperref}

\usepackage[accepted]{icml2024}

\usepackage{amsmath}
\usepackage{amssymb}
\usepackage{mathtools}
\usepackage{amsthm}
\usepackage{wrapfig}

\usepackage{longtable}

\usepackage[capitalize,noabbrev]{cleveref}

\theoremstyle{plain}

\theoremstyle{definition}

\theoremstyle{remark}

\usepackage[textsize=tiny]{todonotes}

\usepackage{tabularx}
\usepackage[toc,page,header]{appendix}
\usepackage{minitoc}

\icmltitlerunning{Case-based or Rule-based: How Do Transformers Do the Math?}

\begin{document}

\twocolumn[
\icmltitle{Case-Based or Rule-Based: How Do Transformers Do the Math?}



\icmlsetsymbol{equal}{*}

\begin{icmlauthorlist}
\icmlauthor{Yi Hu}{PKU}
\icmlauthor{Xiaojuan Tang}{PKU,BIGAI}
\icmlauthor{Haotong Yang}{PKU}
\icmlauthor{Muhan Zhang}{PKU,BIGAI}
\end{icmlauthorlist}

\icmlaffiliation{PKU}{Institute for
Artificial Intelligence, Peking University}
\icmlaffiliation{BIGAI}{National Key Laboratory of General Artificial Intelligence, BIGAI}

\icmlcorrespondingauthor{Muhan Zhang}{muhan@pku.edu.cn}

\icmlkeywords{Machine Learning, ICML}

\vskip 0.3in
]



\printAffiliationsAndNotice{}  

\begin{abstract}
Despite the impressive performance in a variety of complex tasks, modern large language models (LLMs) still have trouble dealing with some math problems that are simple and intuitive for humans, such as addition. While we can easily learn basic \textit{rules} of addition and apply them to new problems of any length, LLMs struggle to do the same. Instead, they may rely on similar \textit{cases} seen in the training corpus for help. We define these two different reasoning mechanisms as ``\textit{rule-based reasoning}'' and ``\textit{case-based reasoning}''. Since rule-based reasoning is essential for acquiring systematic generalization ability, we aim to explore exactly whether transformers use rule-based or case-based reasoning for math problems. Through carefully designed intervention experiments on five math tasks, we confirm that transformers are performing case-based reasoning, no matter whether scratchpad is used, which aligns with the previous observations that transformers use subgraph matching/shortcut learning to reason. 
To mitigate such problems, we propose a Rule-Following Fine-Tuning (RFFT) technique to teach transformers to perform rule-based reasoning. Specifically, we provide explicit rules in the input and then instruct transformers to recite and follow the rules step by step. Through RFFT, we successfully enable LLMs fine-tuned on 1-5 digit addition to generalize to up to 12-digit addition with over 95\% accuracy, which is over 40\% higher than scratchpad. The significant improvement demonstrates that teaching LLMs to use rules explicitly helps them learn rule-based reasoning and generalize better in length. Code is available at \url{https://github.com/GraphPKU/Case_or_Rule}.


\end{abstract}

\section{Introduction}
\label{introduction}
Large language models (LLMs) such as ChatGPT~\citep{chatgpt} and GPT-4~\citep{gpt4} have exhibited remarkable capabilities in a wide range of tasks from some classical NLP tasks such as translation and summarization to complex reasoning tasks about commonsense, math, logic and so on~\citep{chatgpt, gpt4, gpt3, llama, chowdhery2023palm, thoppilan2022lamda}. Some people believe LLMs present a seemingly promising route to AGI~\citep{sparks}. At the same time, some theoretical work also gives their support to this applauded prospect by proving that transformer-based LLMs can learn an intrinsic mechanism for some complex tasks such as linear regression~\citep{akyürek2023linear-model}, dynamic programming~\citep{feng2023revealing} or modular addition~\citep{zhong2023pizza,nanda2023progress,power2022grokking,liu2022understanding}. 

Although LLMs have demonstrated impressive results and possibility both in performance and theory, they are, surprisingly, still puzzled by some basic calculation tasks~\citep{general-purpose, inexperienced_chatgpt, koralus2023humans, dziri2023faith, xu2023mysterious-drop, zhou2023length}. Notably, there has been a line of work paying efforts to teach transformers to perform \textit{addition of two large numbers}~\citep{nye2021scratchpad,qian2022limitations,zhou2022teaching,zhou2023length,shen2023positional,kazemnejad2023impact,lee2023teaching,zhou2024transformers}. Despite ongoing efforts, transformers have yet to successfully generalize to new inputs that are significantly longer than the training data, without relying on external tools. In contrast, humans can easily solve addition of two numbers of any length after learning basic rules of column addition.
Language models often astonish us with their proficiency in complex tasks, yet they can also perplex us with unexpected failures in seemingly straightforward tasks. This dichotomy in performance raises intriguing questions about their underlying reasoning mechanisms. 

Previous work has argued over the open questions. \citet{nanda2023progress, zhong2023pizza} study how transformers do modular addition and claim that they derive certain algorithms to solve the problem, such as the clock algorithm where input numbers are represented as angles and then added together. However, another line of work~\citep{dziri2023faith, wu2023reasoning-or-reciting, zhang2023counterfactual} worries that the impressive reasoning ability of LLMs can be mainly attributed to the extensive training corpus. They argue that transformers are just recalling similar instances from seen data to solve reasoning tasks instead of capturing underlying rules and applying them to new problems. 

\begin{figure}[t]
    \centering
    \includegraphics[width=.5\textwidth]{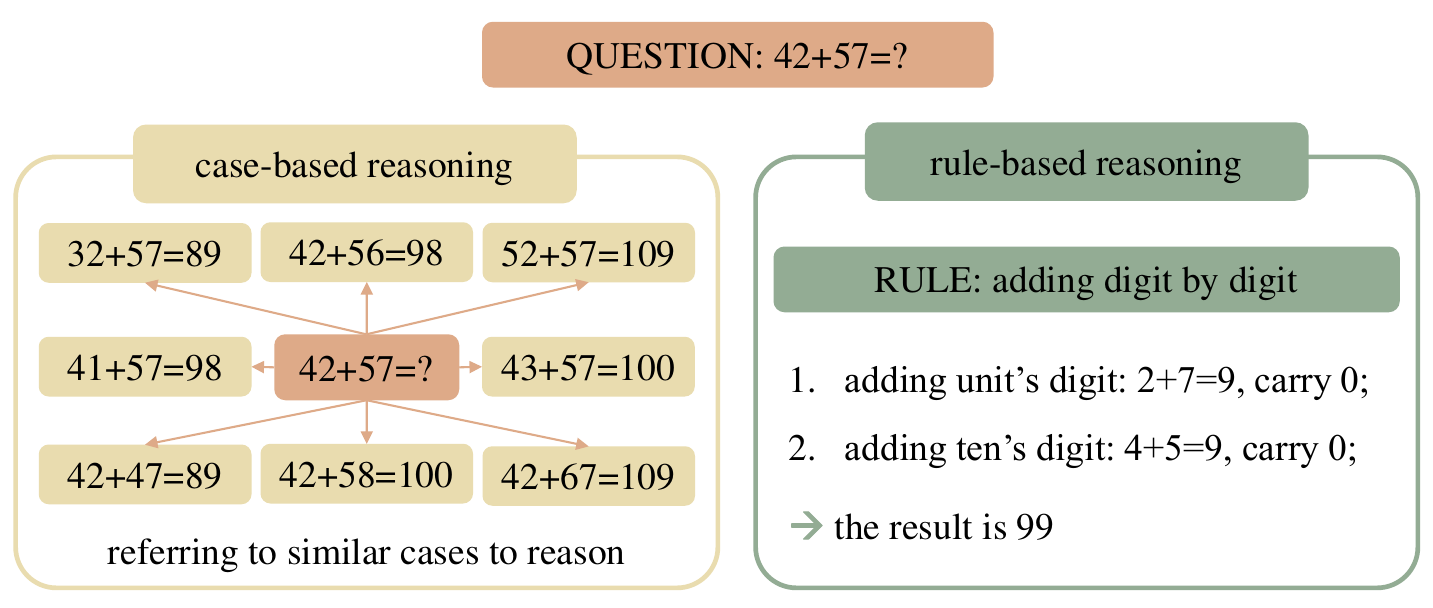}
    \caption{Illustrations of case-based and rule-based reasoning.}
    \label{case-or-rule}
\end{figure}

In this paper, we study the hotly-debated questions more directly through intervention experiments. We hypothesize that transformers significantly depend on certain cases in training data to do math reasoning, which we denote as ``case-based reasoning''. It should be noted that here by ``case-based reasoning'' we do not mean a non-parameterized machine learning algorithm that really retrieves similar cases from a database. Rather, we describe a behavior that transformers show in reasoning. Specifically, if a model employs case-based reasoning, the removal of those dependent cases from the training set would significantly affect its accuracy on certain test examples. On the contrary, if a model does not rely on similar cases but instead masters the underlying \textit{rules} for reasoning---a mechanism we define as ``rule-based reasoning''---the absence of these cases should not affect the performance. An illustration of these two contrasting reasoning paradigms is shown in Figure~\ref{case-or-rule}.

To verify our hypothesis, we fine-tune LLMs respectively on five basic and representative math tasks: addition, modular addition, base addition, linear regression, and chicken \& rabbit problems. As a sanity check for each task, we first make sure that the model achieves 100\% performance on the test set when the dataset is randomly split. Then, we artificially split the dataset by leaving out some continuous regions of examples as the test set with the remaining ones as the training set and re-train the model. This method ensures that most test examples do not have close training cases to support their inference. Our results show that in all tasks, the model performance drops significantly in the second setting, despite that the size of the training set (above 95\% of the whole dataset) is entirely sufficient to achieve 100\% accuracy under random split. See Figure~\ref{fig:holes} and Figure~\ref{fig:3holes} for example. 

The results of our intervention experiments provide direct evidence suggesting that transformers perform case-based reasoning for math problems. This also aligns with previous work~\citep{dziri2023faith} showing transformers rely on seen computation subgraphs for multi-step reasoning. However, there are notable distinctions in our approach and findings: 
\citet{dziri2023faith} look at the frequency difference of seen subgraphs in correct and incorrect samples respectively as indirect evidence that models rely on seen subgraphs to generate correct answers, while we present direct evidence of case-based reasoning by showing the performance gap before and after removing the cases. Besides, we study both single-step and multi-step reasoning while \citet{dziri2023faith} mainly focus on compositional reasoning. 

So why is rule-based reasoning so important? Rule-based reasoning is essential for models to achieve systematic and length generalization so that they can be applied to new, unseen scenarios without re-training.
As our last contribution, we propose a method to shift transformers from case-based to rule-based reasoning, thereby fostering a more robust and generalizable reasoning paradigm. Focusing again on the addition of large numbers, we propose a technique that teaches transformers to follow rules step by step. Specifically, we explicitly put rules in the input and enforce the model to step-by-step recite and follow the necessary rules to complete reasoning, which we call Rule-Following Fine-Tuning (RFFT). 
Through RFFT, LLMs trained on addition of numbers of 1-5 digits successfully generalize to up to 12-digit addition, verifying its effectiveness in teaching LLMs to perform rule-based reasoning. It is noteworthy that the training set is as small as 100 samples, demonstrating that RFFT enables models with sufficient fundamental capabilities to grasp the rules through a small set of examples, which aligns with humans' few-shot rule learning ability.



\section{Related Work}
\label{related_work}
\paragraph{LLM reasoning.} Recent years have seen enormous improvement in LLMs' capabilities. LLMs show impressive performance in a wide range of tasks~\citep{chatgpt, gpt4, gpt3, llama, chowdhery2023palm, thoppilan2022lamda}. However, various tasks of complex reasoning are still challenging for LLMs~\citep{srivastava2022beyond}. In particular,~\citet{dziri2023faith, xu2023mysterious-drop, zhou2023length, zhou2024transformers} show that LMs still struggle with math reasoning, even with basic calculation operations. 

Previous work has come up with methods to simplify the tasks by decomposing them to simpler intermediate steps. For example, ~\citet{nye2021scratchpad, zhou2022teaching} introduce finetuning models with cases containing scratchpads to improve arithmetic reasoning of LLMs.~\citet{few-shot-cot, zero-shot-cot, zhou2023leasttomost, khot2023decomposed, zhu2023large} propose various prompting methods to teach the model to generate rationales before the final answer with in-context learning. However, even with these methods, LLMs are still far from completely solving arithmetic reasoning tasks. The failures inspire us to study how exactly LLMs perform math reasoning. Besides, we study the effects of the methods of task simplification on case-based reasoning in our paper. Specially,~\citet{zhu2023large} improve the model performance by providing the cases the reasoning process may depend on in the input, which in fact aligns with our case-based reasoning paradigm.

\paragraph{Memorization or generalization.} As reasoning capabilities of LLMs can be mainly attributed to the scaling effects of the training corpus and the model size, the question of whether the seemingly impressive reasoning abilities are the results of capturing general rules lying under the natural language or just reciting seen cases from the huge training corpus is drawing more and more attention. \citet{wu2023reasoning-or-reciting, zhang2023counterfactual} investigate into the gap of capabilities of LLMs to conduct reasoning over factual and counterfactual situations and show the significant performance drop in counterfactual cases, suggesting LLMs are reciting answers of common cases in the training corpus. A recent work~\citet{dziri2023faith} models reasoning tasks as computation graphs and show empirically that LLMs conduct reasoning via subgraph matching instead of developing systematic problem-solving skills. We study the question of interest in a straightforward way by removing certain samples from the training set and show significant performance gap. By tracing back to the effective training datapoints, we confirm that transformer-based LLMs are relying on surrounding cases in the training set to do math reasoning instead of learning generalizable rules.
On the other hand, \citet{hou2023mechanistic} study the problem through probing the models' attention patterns and claim that transformers are implementing reasoning trees in the reasoning process.
\citet{yang2023explaining} propose that LLM's reasoning ability comes from memorizing some templates, which are some fixed parts in the reasoning process, enabling generalization within tasks.

\paragraph{Grokking.} Recent work has shown the phenomenon of model capturing generalizable rules of arithmetic reasoning tasks long after overfitting the training set, known as grokking~\citep{power2022grokking, liu2022understanding}.~\citet{nanda2023progress, zhong2023pizza} study the algorithms transformers learn in the task of modular addition. The series of work show through experiments that the model learns systematic rules to solve modular addition through embedding the numbers as angles and operating on their trigonometric functions. We also try to observe the phenomenon in the same setting as in~\citet{zhong2023pizza} with certain samples removed from the training set.
Although we observe the growth of test performance after the model overfitting the training set, there is still a wide gap between training accuracy and test accuracy, suggesting the model fails to learn the rules. This phenomenon indicates that even the ability to learn and apply generalizable arithmetic algorithms in grokking deeply depends on certain cases in the training set. The results and experiments are described in Appendix~\ref{grokking}.

\paragraph{Theoretical expressiveness.} ~\citep{feng2023revealing,akyürek2023linear-model,dai2023gradient-descent,vonoswald2023gradient-descent,garg2023simple-function}
There have been a large number of work studying the expressive power of transformers.~\citet{yun2020universal} proved that transformers are universal approximators of continuous sequence-to-sequence functions on a compact domain. More recently,~\citet{garg2023simple-function} reveals that auto-regressive transformers can learn basic functions including sparse linear functions, MLPs and decision trees. Furthermore,~\citet{akyürek2023linear-model} demonstrates that transformers can in-context learn linear regression by implementing the algorithm of gradient descent~\citep{dai2023gradient-descent,vonoswald2023gradient-descent}.~\citet{feng2023revealing} shows how chain-of-thought prompting help transformers complete tasks including basic calculations, linear equations and dynamic programming. In our work, we conduct empirical experiments and show how auto-regressive transformers do basic math reasoning in practice. We include tasks like addition, linear regression and linear functions that have been studied in the theoretical work.

\paragraph{Length generalization.}
Length generalization calls for the ability to generalize to longer sequences than seen in training samples, which remains a challenge for transformers~\citep{abbe2023generalization, anil2022exploring, zhou2023length}. Previous work has shown that data format and positional encoding are crucial to length generalization ability through experiments on small transformers across various tasks such as arithmetic reasoning~\citep{lee2023teaching, kazemnejad2023impact, shen2023positional, zhou2023length, zhou2024transformers}. However, these works require specifically designed tricks for each task and train small transformers from scratch. Our work explores length generalization in the settings of fine-tuning pre-trained LLMs and shows that the technique of RFFT we propose in \S\ref{sec:experiment_rule} greatly enhances length generalization. Furthermore, we demonstrate that the models with sufficient fundamental capabilities can generalize well with only a small set of training samples.

\section{Case-based and Rule-based Reasoning}
\label{sec:case-based_or_rule-based}
One main focus of our paper is to discuss whether auto-regressive transformer-based language models are solving basic math problems based on cases or rules. In this section, we intuitively motivate these two reasoning paradigms and provide a direct method to distinguish them through data intervention. 


\paragraph{Case-based Reasoning.} A model engaging in case-based reasoning exhibits sensitivity in its test performance to the division of the dataset into training and test sets. Specifically, if a model relies on shortcuts, either by referencing similar cases encountered during training or by merely repeating previously seen examples to solve new problems, its effectiveness diminishes when these cases are removed from the training set. This reduction in relevant training data results in a notable decrease in the model's ability to accurately respond to test questions.

\paragraph{Rule-based Reasoning. } In contrast to case-based reasoning, the paradigm of rule-based reasoning allows the model to learn the underlying rules, which are insensitive to the data split. For example, if a model is developing the systematic rules of addition during the training process, its test performance should not be affected severely if we leave some of the training samples out of the training set and add some others to keep the same training-test ratio. It should be noted that the training set should always provide the necessities for the model to learn the underlying rule. For example, the training set should at least cover all the tokens used in the test set in order to develop a systematic rule that applies to the whole dataset. In all our experiments, we carefully design the setups to ensure the above.

Based on the above discrimination, we propose a natural method to determine whether a model is performing case-based reasoning or rule-based reasoning through data intervention. That is, we artificially remove certain regions of the training data to see its effect on test performance. For example, in math reasoning tasks such as addition, if the model is severely relying on some seen cases to do reasoning, a natural hypothesis is that it is relying on some surrounding cases of the test question, as shown in Figure~\ref{case-or-rule} left. Based on the hypothesis, we can remove a small set of surrounding cases from the training set and see whether the model can still answer the question. If it succeeds when we leave the surrounding cases in the training set but fails when we take them out, we can judge that the model is relying on the small set of surrounding cases to do math reasoning. Otherwise, if the model can perform well in the test set no matter how we split the dataset, it is likely performing rule-based reasoning which guarantees robust generalization independent of dataset split.


It is important to recognize that rule-based reasoning also involves a degree of memorization. For example, in the process of digit-by-digit addition, we inherently rely on memorized knowledge of possible single-digit sums. Take the calculation of 42+57 as an instance; it is essential to know that 2+7 equals 9 and 4+5 equals 9. We refer to this fundamental knowledge required for rule-based reasoning as ``unit rules''. These unit rules are tailored to specific reasoning patterns. The more basic these unit rules are, the less memorization the reasoning process requires, indicating a more pronounced reliance on rule-based reasoning. Conversely, if a model relies on case-based reasoning through sheer memorization---learning that 42+57 equals 99 only by encountering this exact case, then the unit rules for this pattern of reasoning are the cases themselves.

So how do we judge whether the unit rules are elemental enough to ensure a rule-based rather than case-based reasoning? We define the model is performing rule-based reasoning if the set of unit rules the model requires to solve the task is \textbf{finite} and can be easily covered by a training set of a reasonable size. Otherwise, if it is hard or even impossible for a training set to cover all the unit rules, we consider the model performing case-based reasoning.

\begin{figure*}[t]
    \centering
    \includegraphics[width=.9\textwidth]{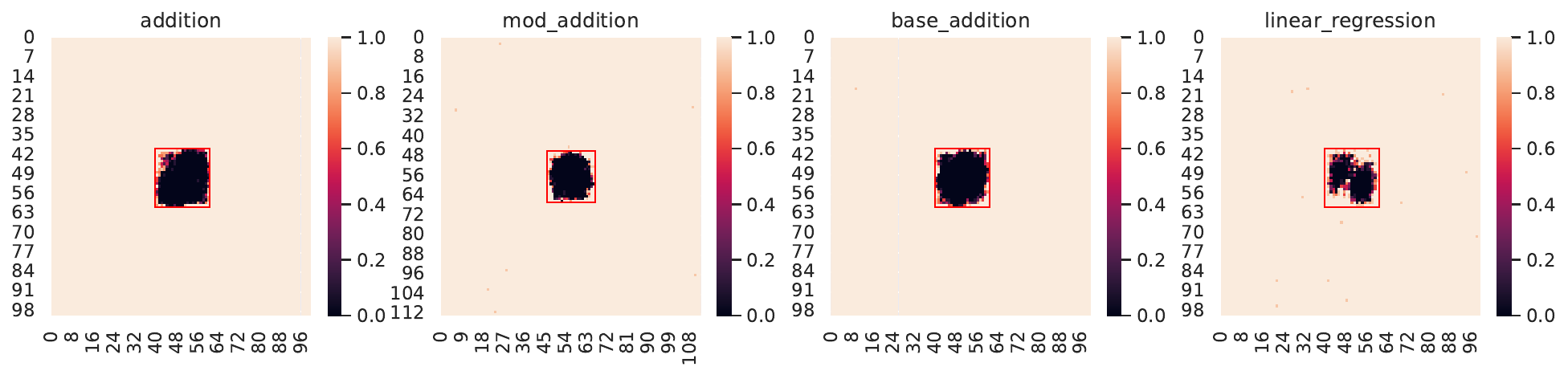}
    \vspace{-10pt}
    \caption{Accuracy of Leave-Square-Out method on addition, modular addition, base addition, and linear regression. The vertical and horizontal axes are $a$ and $b$, respectively. The area inside red boxes represents the test squares. During generation, we set the model temperature to 1 and sample 10 generations to evaluate the accuracy on each test point. We only leave one test square out in this experiment. The square center \((a_k, b_k)\) is (50, 50) for addition, base addition and linear regression and (56, 56) for modular addition.}
    \label{fig:holes}
\end{figure*}

\begin{figure*}[t]
    \centering
    \includegraphics[width=.9\textwidth]{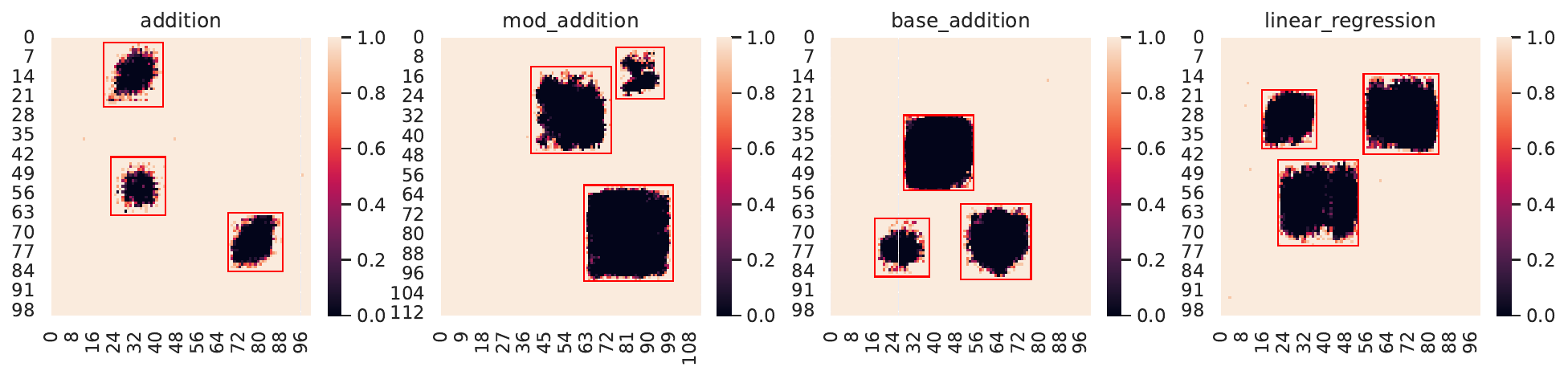}
    \vspace{-10pt}
    \caption{We randomly select 3 centers of test squares $(a_k, b_k)$ and corresponding lengths $l_k$ ranging from 20 to 40 to see whether the locations and the side lengths affect the case-based reasoning behavior for datasets including addition, modular addition, base addition and linear regression. The area inside red boxes represents the test squares. We sample 10 generations at each data point and report the accuracy. The figure shows that holes consistently appear with the locations and side lengths of the test squares varying.}
    \label{fig:3holes}
\end{figure*}

\section{Transformers are Doing Case-based Reasoning}
\label{sec:experiment_case}
In this section, we provide direct evidence that transformers perform case-based reasoning through intervention experiments on five representative math tasks.
\subsection{Experimental Setup}
\paragraph{Datasets}
We focus on binary operations, which take two numbers \(a, b\) as inputs. Denoting $c$ as the target label, we construct datasets like \(\mathcal{D}=\{((a_i, b_i), c_i)\}\) for five math tasks including addition, modular addition, base addition, linear regression, and chicken \& rabbit problem:
\begin{itemize}[itemsep=2pt,topsep=-2pt,parsep=0pt,leftmargin=10pt]
\item \textbf{Addition.} The input to the transformer is ``\(a+b\)'', the output is ``\(c\)'', where \(c=a+b\). \(a, b\) range from \(0\) to \(99\).

\item \textbf{Modular addition.}  The input to the transformer is ``\(a+b\)'', the output is ``\(c\)'', where \(c=a+b\mod P\).  \(a, b\) range from \(0\) to \(112\). We set \(P=113\) as a constant.

\item \textbf{Base addition.}  
This task is the same as addition, except that all numbers $a,b,c$ are expressed in the base-$n$ numerical system. In this paper, we set \(n=9\) as a constant.

\item \textbf{Linear regression.} This task requires the transformer to learn a linear regression function. The input is ``\((a,b)=\)'', the output is ``\(c\)'', where \(c=m\cdot a+n\cdot b+p\).  \(a, b\) range from \(0\) to \(99\). We set \(m=1, n=2, p=3\) as constants.

\item \textbf{Chicken \& rabbit problem.} We construct a dataset of chicken \& rabbit problems with natural language questions and answers. The input to the transformer is ``Q: Rabbits have 4 legs and 1 head. Chickens have 2 legs and 1 head. There are \(a\) legs and \(b\) heads on the farm. How many rabbits and chickens are there?''. The output is ``A: There are \(c\) rabbits and \(d\) chickens.'', where \(c=(a-2b)/2, d=(4b-a)/2\). \(b\) ranges from \(0\) to \(99\). For each \(b\), \(a\) ranges from \(2b\) to \(4b\) with a step of 2. It is a representative task involving solving a system of linear equations. 
\end{itemize}
\paragraph{Models}
We use GPT-2, GPT-2-medium~\citep{gpt2}, and Llama-2-7B~\citep{llama2} in this section. We fine-tune GPT-2 by default respectively on each dataset for 100 epochs under different training-test splits, with batch size set to 30 and learning rate set to \(10^{-4}\). 

\subsection{Method}
\label{method}
\paragraph{Leave-Square-Out}
 To test whether the model is relying on certain cases to solve the problem, we need to first locate such cases and then remove them from the training set to see whether they affect the model performance. Our hypothesis is that when facing a certain test sample, transformers tend to rely on training samples ``close'' to the test sample to perform reasoning. Thus, we construct a square test set to \textbf{isolate the test samples from the training samples}. For example, suppose the square center is \((a_k, b_k)\) and the side length is \(l_k\), we construct a square test set as \(\mathcal{T}_k=\{((a_i, b_i), c_i)~|~ a_k - \frac{l_k}{2} \leq a_i \leq a_k + \frac{l_k}{2}, b_k - \frac{l_k}{2} \leq b_i \leq b_k + \frac{b_k}{2}\}\). All the remaining samples constitute the training set. According to our hypothesis, case-based models should fail to generate correct answers for test samples near \((a_k, b_k)\), as there are no close cases in the training set.

\subsection{Appearance of Holes Verifies Case-Based Reasoning}
In our study, we apply the Leave-Square-Out method to each dataset. Specifically, we extract a square comprising 441 samples (from a total of approximately 10,000 samples) with a side length of 20 to form our test set, leaving the remainder as the training set. It is important to note that, despite removing a small portion of training samples, we ensure that all tokens present in the dataset appear in the training set. This precaution is to prevent the models from failing simply due to encountering unseen tokens. We then proceed to fine-tune GPT-2 and GPT-2-medium models using this specific training-test split for each dataset. For comparison, we also fine-tune these models on datasets that are randomly split, where each training set comprises 70\% of the total dataset.

Models achieve 100\% accuracy easily in the random split settings across all datasets, which suggests that the size of training sets in the Leave-Square-Out setting (above \textbf{95\%} of each dataset) is totally sufficient to complete the task. However, in the Leave-Square-Out setting, as shown in Figure~\ref{fig:holes}, there are ``holes'' appearing in the accuracy distribution of the test squares over $a$ and $b$. The appearance of holes in the figure indicates that the test samples away from the boundary of the training set are hard for the models to correctly infer, while the models can easily handle the test samples near the boundary. This suggests that in the basic math reasoning tasks, when faced with an unseen test case, transformers \textbf{rely on the surrounding training cases} to predict the answer, verifying the case-based reasoning hypothesis. As for random split, every test sample has close training samples to support its inference, thus reaching 100\% accuracy. 
In Figure~\ref{fig:holes}, we only show the results of GPT-2 on the first four tasks; the results of GPT-2-medium and the results of chicken \& rabbit problem are shown in Appendix~\ref{sec:additional_results}.

\begin{figure*}[t]
    \centering
    \includegraphics[width=.9\textwidth]{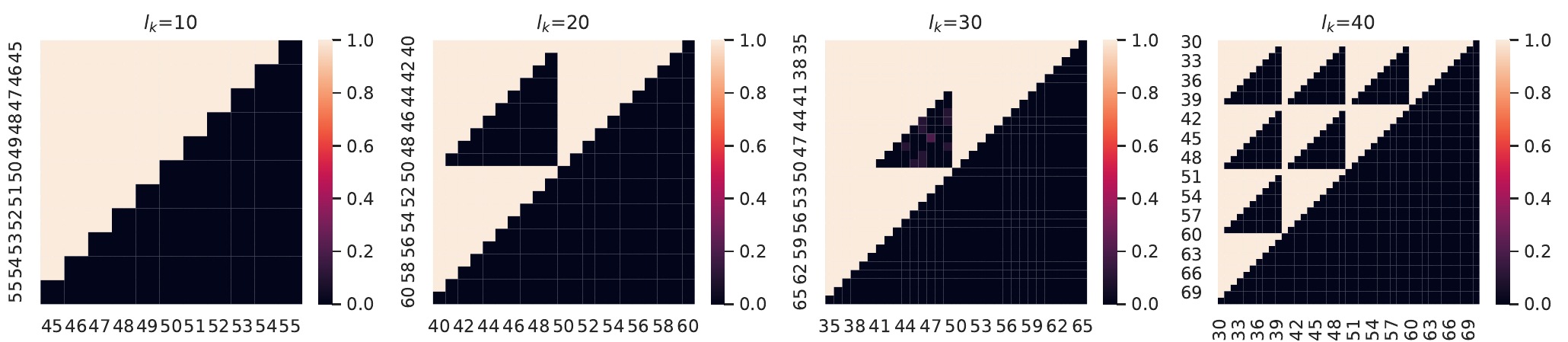}
    \vspace{-5pt}
    \caption{Test accuracy distribution of GPT-2 trained with scratchpad in the task of addition. Note that all points in the figure are test samples; each subfigure here corresponds to a left-out square in the original plane. From left to right, the side length of test square is set to $l_k=10, 20, 30, 40$. For each test point, we sample 10 generations and show the accuracy of generating the correct answer.}
    \label{scratch}
\end{figure*}

\begin{figure}[t]
    \centering
    \includegraphics[width=.4\textwidth]{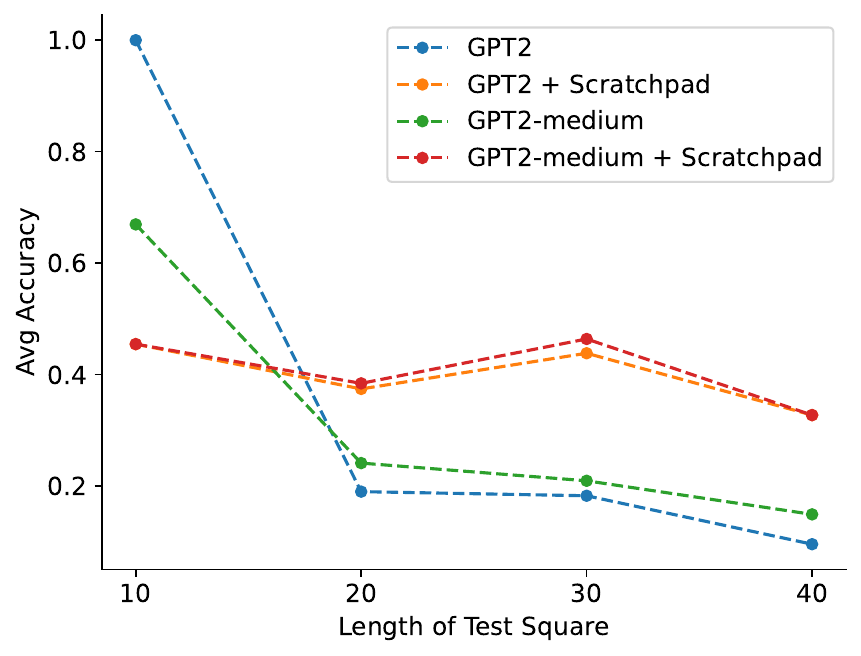}
    \caption{In the task of addition, we show the average accuracy over all test samples (samples within the square) with side length $l_k=10, 20, 30, 40$. We test four models: GPT-2, GPT-2 with scratchpad, GPT-2-medium and GPT-2-medium with scratchpad.}
    \label{fig:ablation}
\end{figure}

\subsubsection{Do locations size of test squares matter?}
To see whether the locations of test squares affect the experimental results, we randomly select three square centers $(a_k, b_k)$ and three corresponding side lengths $l_k$ ranging from 20 to 40 for each dataset. As shown in Figure~\ref{fig:3holes}, holes consistently appear in various locations of the dataset, suggesting that the model behavior of performing case-based reasoning does not change with the location of test sets.

\subsubsection{Does the size of test square matter?}
We test on various lengths of test squares including $l_k$ set to 10, 20, 30, 40 with GPT-2 and GPT2-medium. As shown in Figure~\ref{fig:ablation}, test accuracy drops with the side length of the test square increasing. This phenomenon is natural in the context of case-based reasoning. As the test square becomes larger, the ratio of test samples that do not have close supporting training samples becomes higher, thus decreasing the test accuracy. Besides, it is shown in Figure~\ref{fig:ablation} that GPT-2 achieves 100\% accuracy when we set $l_k$ to 10. In other words, \textbf{the hole disappears when the test square shrinks to less than a small size} where all the samples in the test set have close training samples for the model to refer to.

\subsubsection{Does adding scratchpad help?}
\label{scratchpad}
\begin{figure*}[t]
    \centering
    \includegraphics[width=.9\textwidth]{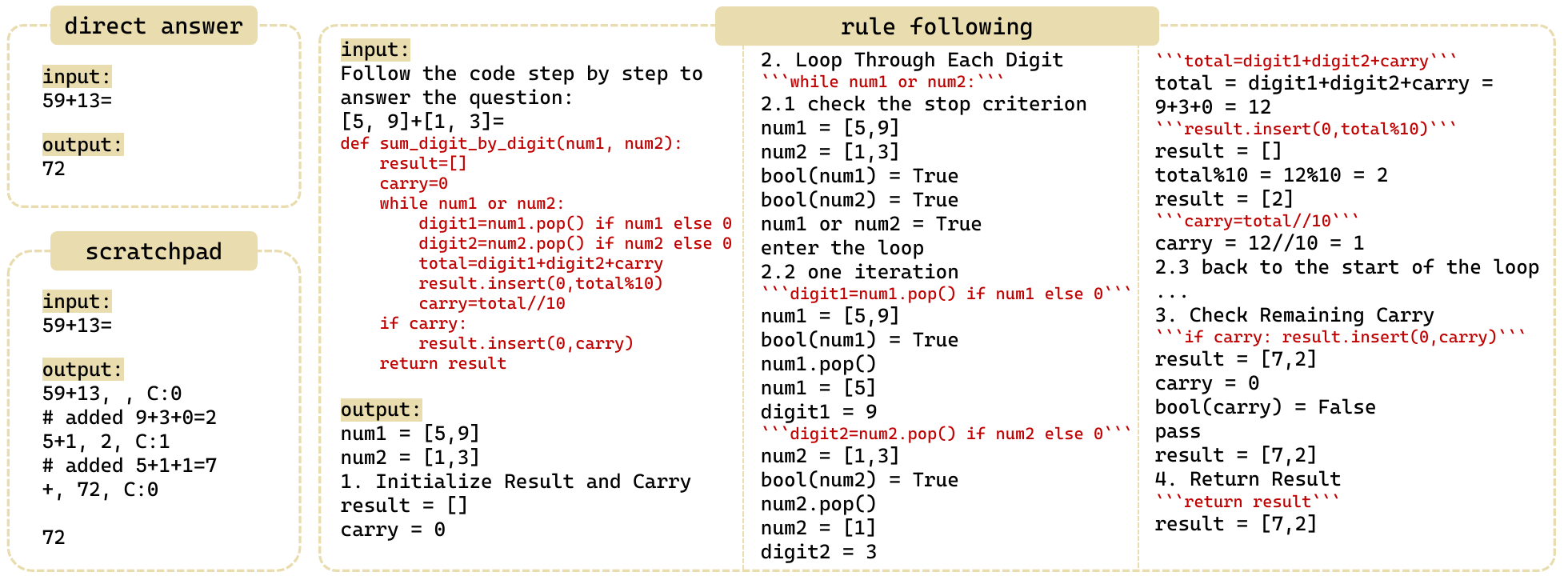}
    \vspace{-5pt}
    \caption{Examples of input-output sequence of question $59+13$ in 3 different settings, including direct answer, scratchpad and rule following. In the setting of rule-following, we provide the Python program of adding two numbers together digit by digit in the input, and provide the step-by-step rule-following process in the output. Examples of the full input-output pairs are shown in Appendix~\ref{full_IO}.}
    \label{prompt}
\end{figure*}

\citet{nye2021scratchpad} has proposed a technique of teaching models to explicitly generate intermediate computation steps into a ``scratchpad'' before arriving at the final answer to improve their math reasoning capabilities. The scratchpad technique enables the model to decompose addition into incremental digit-by-digit operations, potentially reducing the model's dependence on surrounding cases. An example input-output pair of scratchpad is shown in the bottom left of Figure~\ref{prompt} (scratchpad). We employ scratchpad fine-tuning to examine its impact on the model's tendency towards case-based reasoning, specifically investigating whether the scratchpad technique can enable transformers to perform rule-based reasoning.

In particular, we alter the input of the addition dataset by providing scratchpad steps of adding two numbers digit by digit before presenting the final answer, instead of directly providing the answer following the question. Then we perform Leave-Square-Out on the altered dataset with GPT-2 and GPT-2-medium. The test accuracy vs. side length results are also shown in Figure~\ref{fig:ablation}. In the settings where side length of the left-out square $l_k\geq 20$, adding the scratchpad greatly boosts the model performance. However, for $l_k=10$, models trained with scratchpad inputs lag behind those trained with direct answers. Besides, the test accuracy of models trained with scratchpad maintain relatively stable with the increase of test square's side length, in contrast to the sharp decline in performance seen in models trained with direct answers. 
To explain the phenomenon, we show the test accuracy distribution over $a$ and $b$ of models trained with scratchpad in Figure~\ref{scratch}. It is clear that the model behavior of relying on cases ``nearby'' to solve new problems has changed. The holes shift to (a series of) triangles with their hypotenuses along the ``carry boundary'' at the unit’s and ten’s digits. %
For example, in the setting of $l_k=20$ (the second subfigure), there are two triangle holes where the model shows almost zero accuracy. We explain why the model fails in each triangle and why the model succeeds in the rest of the test set as follows. Firstly for the small triangle, the model fails to answer questions like 47+48. 47+48 can be decomposed into 2 steps: 7+8=5, carry 1; 4+4+1=9. As \textbf{there are no cases in the training set containing the step of 4+4+1 in the ten’s digit, the model fails}. In contrast, for those test points that do not involve carry in the ten's digit, like 42+43, the model succeeds because it can learn 4+4=8 from plenty of training data.  Secondly for the large triangle, the model fails to answer 57+58. 57+58 can be decomposed into 2 steps: 7+8=5, carry 1; 5+5+1=1, carry 1. As there are no training cases performing 5+5 in the ten’s digit which requires carry 1 to the hundred’s digit, the model fails. 

The shapes and locations of the holes indicate that the models succeed in test cases where \textbf{every step of the corresponding scratchpad has appeared in the training set} and fail otherwise. This conclusion aligns with \citet{dziri2023faith} that transformers rely on seen computation subgraphs for complex reasoning. More importantly, this phenomenon demonstrates even \textbf{scratchpad cannot teach transformers to perform rule-based reasoning}---the models still \textbf{mechanically recite the seen unit rules}, but fail to flexibly generalize them.

\subsubsection{Does the model and data size matter?}
As the emergent ability~\citep{wei2022emergent} suggests that the model size is crucial to unlocking a wide range of complex tasks, we first explore the effects of model size on the reasoning mechanisms through experiments on GPT-2, GPT-2-medium and Llama-2-7B. GPT-2 has 124M parameters, and GPT-2-medium has 355M parameters. As shown in Figure~\ref{fig:ablation}, when trained with direct answers instead of scratchpads, GPT-2-medium generally outperforms GPT-2 when $l_k\geq 20$. On the contrary, GPT-2-medium lags behind GPT-2 when $l_k=10$. Besides, when trained with scratchpad, GPT-2-medium performs slightly better than GPT-2. Overall, model size has a more pronounced impact on test performance in scenarios of training with direct answers, as opposed to training with scratchpad, probably because the single steps in scratchpad is easier to memorize.
We put the experiments on Llama-2-7B and GPT-3.5 to Appendix~\ref{app:model_size_llama} and Appendix~\ref{app: model_size_gpt3.5}. Both models show holes within the test square, indicating that the trend of case-based reasoning still exists.

We also study how data size affects the behavior of case-based reasoning. We expand the range of $a, b$ from 100 to 200 and 500, respectively. We also scale up the side length of the test square linearly with the data range. With the increasing of data size, the holes still appear, suggesting that increasing the data size helps little. We show the test accuracy distribution in Appendix~\ref{app:data_size}.

\subsection{In-context Learning}
\label{sec:case_icl}
Another aspect of LLMs' reasoning ability is attributed to in-context learning (ICL). This method draws upon knowledge not only ingrained during the pre-training but also from specific examples supplied within the context. The underlying mechanisms that make ICL effective are among the most intriguing and unanswered questions in the field. We extend our investigations to ICL in Appendix~\ref{sec:ICL}, revealing that LLMs' ICL reasoning ability also exhibits characteristics of case-based learning.

\section{Teaching Transformers to Do Rule-Based Reasoning by Rule-Following Fine-Tuning}
\label{sec:experiment_rule}
In~\S\ref{sec:experiment_case}, we show that transformers are performing case-based reasoning in a wide range of math problems. However, the case-based reasoning behavior sets strong limits to the generalization ability of transformers. To be more specific, based on the results in~\S\ref{sec:experiment_case}, transformers rely on surrounding cases to do addition, so they naturally cannot generalize in length by training on finite-digit addition data. In contrast, rule-based reasoning can robustly generalize in length. In this section, we explore how to teach transformers to do rule-based reasoning. 

We first revisit the failure of the scratchpad attempt. Despite providing step-by-step intermediate computations, scratchpad fine-tuning fails to teach transformers the \textbf{actually applied ``rule''} behind each step. This is like teaching children addition \textbf{only by showing them examples}, without telling them the \textbf{rationales behind each step}. Motivated by this intuition, we propose Rule-Following Fine-Tuning (RFFT) to explicitly teach transformers to use rules at each step.

RFFT has two steps. First, we explicitly list the rules for solving a given task in the input. For example, in the task of addition, we provide the code of adding two long integers digit by digit in the input. It should be noted that there are various ways to represent the rules, including programs, pseudo-code, first-order logic, natural language, etc. We use programs in this section, and explore using natural language representations of rules in Appendix~\ref{app:code_nl}. 
Second, we fine-tune the model to follow the rules step by step. Specifically, the model need to explicitly recite which rule it is using in each step, as well as updating the intermediate variables after applying this rule, as shown in Figure~\ref{prompt} right.



\subsection{Experimental Setup}\label{sec:RFFT_exp}
In this section, we use two models, Llama-2-7B and GPT-3.5-turbo-1106. We fine-tune Llama-2-7B ourselves, and fine-tune GPT-3.5-turbo-1106 through the OpenAI API service.
We focus on the length generalization problem of addition of two large numbers $a$ and $b$, and put additional experiments on the task of concatenating last letters to Appendix~\ref{sec:last_letter}. We randomly sample $a$ and $b$ to construct the training data, where the numbers of digits of $a$ and $b$ range from 1 to 5, constituting about 500k samples in total for Llama-2-7B. When fine-tuning GPT-3.5, we reduce the training set to as small as 100 samples. We expect models with sufficient fundamental capabilities to be able to grasp rules through only a small set of training cases, which aligns with how humans learn calculations. During test, we randomly generate 1,500 samples for each digit length from 1 to 9 for Llama-2-7B, and generate 500 samples for each digit length from 6 to 15 for GPT-3.5. The digit length considers the context window size of each model. For GPT-3.5, due to the smaller training set, we perform five independent experiments and report the average accuracy and standard deviation. We employ direct answer, scratchpad, and RFFT as three fine-tuning methods for comparison. 
The training details are shown in Appendix~\ref{rule_detail}.

\begin{figure}[t]
  \centering
  \subfigure[Accuracy of Llama-7B fine-tuned with three methods tested on addition with 1-9 digits.]{
    \label{fig:llama_ft} 
    \includegraphics[width=.35\textwidth]{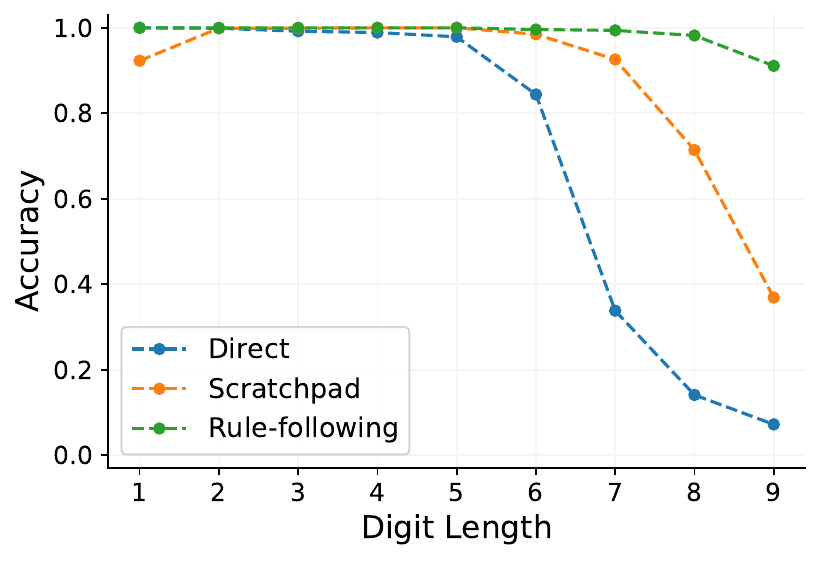}}
  \subfigure[Accuracy of GPT-3.5 fine-tuned with three methods tested on addition with 6-15 digits.]{
    \label{fig:gpt3_ft} 
    \includegraphics[width=.35\textwidth]{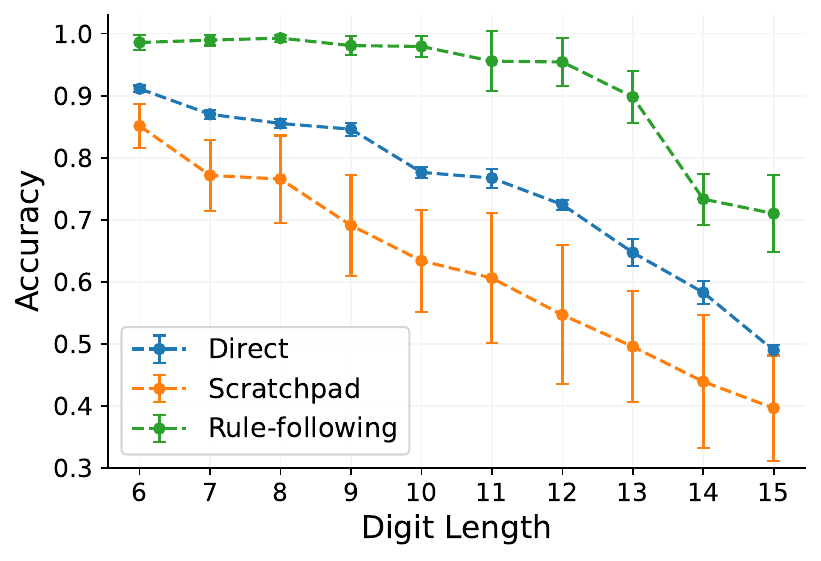}}
  \caption{Accuracy of Llama-2-7B and GPT-3.5-turbo fine-tuned with direct answer, scratchpad and rule following on addition.}
  \label{fig:rfft} 
  \vspace{-5pt}
\end{figure}

\subsection{Results and Analysis}
\label{rule_result}
\paragraph{Overall Results} The results are presented in Figure~\ref{fig:rfft}. Overall, \textit{rule-following} significantly outperforms \textit{direct} and \textit{scratchpad}. When using Llama-2-7B with Rule-Following Fine-Tuning (RFFT), the model shows impressive generalization capabilities in performing addition with 6 to 9 digits, maintaining 91.1\% accuracy even with 9-digit sums.  In comparison, the scratchpad method achieves less than 40\% accuracy in similar tasks. With GPT-3.5-turbo, which possesses more advanced foundational abilities, the RFFT method enables it to astonishingly generalize to additions involving up to 12 digits, with still over 95\% accuracy on 12-digit addition despite seeing only 100 training samples. This significantly surpasses the results from the scratchpad and direct answer fine-tuning methods. 
These results highlight the effectiveness of our Rule-Following Fine-Tuning technique in steering transformers towards rule-based reasoning, showcasing its potential in enhancing model generalization. We provide detailed ablation studies in Appendix~\ref{rfft_ablation}. 

\paragraph{Error Analysis}
We also delve into failure cases to investigate why rule-following fails to achieve a perfect generalization with 100\% accuracy. 
We find that the models can always select the right rule to execute in each step in a recursive way, but sometimes make mistakes when executing some basic operations, such as ``pop''. Consider the example ``\texttt{num2=[9,0,7,6,9,3,7]}''; the expected output after ``\texttt{num2.pop()}'' should 
 be ``\texttt{num2=[9,0,7,6,9,3]}'', while the models in some rare cases will generate ``\texttt{num2=[9,0,7,6,9]}''. As the length increases (e.g., more than 9-digit addition), the phenomenon becomes more severe, which could be attributed to hallucinations or the limited long context abilities of current LLMs~\citep{li2023loogle}. As mentioned in \citet{min2023factscore}, the tendency for hallucinations grows as the length of the generated content expands. These basic capabilities of LLMs might be the bottleneck that limits their strict length generalization under RFFT. It is also analogous to that we humans also tend to make sloppy mistakes when calculating long numbers by copying the wrong digits or forgetting to carry.

\paragraph{Comparison to Scratchpad} Our RFFT technique provides the \textbf{explicit} rules in the input and also teaches LLMs to \textbf{quote} the part of rules used in each step, which helps LLMs understand what each step is doing without having to refer to the long preceding texts.
For example, with clear instructions ``\texttt{total=digit1+digit2+carry}'', an LLM knows it need to find and add these three variables together. In comparison, scratchpad requires LLMs to learn that the third number ``0'' in the formula ``7+6+0=3'' is the carry from last digit, increasing the difficulty of learning. Some example errors of RFFT and scratchpad are included in Appendix~\ref{app:failure_RFFT_scratchpad}. We also discuss RFFT's differences from scratchpad tracing in Appendix~\ref{app:code_nl}.

\paragraph{RFFT as a Meta Learning Ability}

As mentioned in~\S\ref{sec:RFFT_exp}, we find that Llama-2-7B requires 150k training samples to generalize to 9 digits while GPT-3.5 can grasp the rules and generalize to 12 digits with only 100 samples. Thus, we hypothesize that rule-following is a meta learning ability---it might be ``learned'' through pre-training on diverse rule-following data and transfer to new unseen domains, and the stronger the foundation model is, the easier it can understand and learn the rules. This also aligns with human's ability to learn new rules, where experienced learners often learn much faster. To provide more evidence, we further fine-tune a larger model Llama-2-70b than Llama-2-7b and a slightly weaker model davinci-002 than GPT-3.5. Our results show that stronger models indeed need less examples to learn rules. See details in Appendix~\ref{app:RFFT_meta}.

\paragraph{Scratchpad vs Direct Answer}
We observe that GPT-3.5, when fine-tuned with scratchpad, underperforms that with direct answer fine-tuning, which contradicts with our intuition that scratchpad is more suitable for arithmetic tasks as well as the results observed in Llama-2-7B. This phenomenon might be attributed to the different mechanisms of addition between scratchpad and direct answer. For example, scratchpad performs digit-by-digit addition from the lowest digit to the highest one, while direct answer always generates the highest digits first. Fine-tuning with scratchpad would strongly change the inherent addition mechanism of the model. At the same time, integer addition is in fact a relatively familiar task for GPT-3.5, wherein the model exhibits some degree of addition ability even when asked to directly generate the answer with an accuracy of $46.2\%$ on 15-digit addition. This makes adopting scratchpad not always more helpful than direct answer fine-tuning. In contrast, RFFT explicitly interpret the step-by-step mechanism, making learning the addition rules much easier. To further support our hypothesis, we increase the number of training examples for scratchpad to 5,000 and observe much improved performance.
See Appendix~\ref{app: scratch_vs_direct} for details. 

\subsection{In-context Learning}
As we discussed in~\S\ref{sec:case_icl}, LLMs encounter difficulties in autonomously \textit{extracting} rules from ICL examples. The subsequent inquiry pertains to the capacity of LLMs to follow \textbf{explicit} rules supplied by in-context examples. Our conclusion is that given detailed rules, LLMs have certain abilities to follow the rules, which allows the models to show some reasoning ability on unfamiliar tasks. However, they do not gain a competitive edge from the rules in tasks already familiar to them. See Appendix~\ref{app:in_context_learning_rf}.

\section{Conclusion}
In our paper, we study whether transformers are performing ``case-based reasoning’’ or ``rule-based reasoning’’ when solving math problems. First, we describe the two reasoning paradigms and show how to distinguish one from the other. Then, we show through intervention experiments on five basic math tasks that transformers are relying on surrounding cases to do math reasoning. To mitigate the limitations of case-based reasoning, we propose a Rule-Following Fine-Tuning (RFFT) framework to teach transformers to perform rule-based reasoning by asking the model to explicitly quote and follow the rule used in each step. RFFT outperforms scratchpad fine-tuning by large margins, successfully enabling GPT-3.5-turbo fine-tuned on 1-5 digit addition to generalize to up to 12 digit addition.

\newpage
\section*{Acknowledgements}
This work is partially supported by the National Key R\&D Program of China (2022ZD0160300), the National Key R\&D Program of China (2021ZD0114702), the National Natural Science Foundation of China (62276003), and Alibaba Innovative Research Program.

\section*{Impact Statement}
Our work provides a new perspective to understand how LLMs do math reasoning. The research for the first time defines and discriminates the two reasoning paradigms and proposes an effective method to steer transformer to perform rule-based reasoning, which is key to systematic generalization. Our work demonstrates that we can also directly teach rules to LLMs instead of just feeding data examples, just like how we teach children to perform addition. 

\nocite{}

\bibliography{reference}
\bibliographystyle{icml2024}

\newpage
\appendix
\onecolumn

\section{Limitations}
Rule-following fine-tuning (RFFT) is temporarily a task-specific method and requires carefully designed input-output sequences. Besides, we mainly focus on fine-tuning models for specific tasks instead of exploring the reasoning mechanism of pretrained models. There also remain further questions to be explored, for example, how various prompting techniques, such as CoT, least-to-most prompting, and program-aided prompting, affect the model behavior of case-based reasoning or rule-based reasoning. These questions are left for future work.

\section{Collections of Hyper-parameters}
We list all the hyper-parameters used in the paper in Table~\ref{tab:hyperparams}.

\begin{table}[H]
\centering
\begin{tabular}{c|cccc}
\toprule
Models      & training epoch  & batch size & learning rate      \\
\midrule
\multicolumn{4}{l}{\textbf{\textit{case-based reasoning}}}         \\
GPT-2       & $100$           & $30$       & $1\times 10^{-4}$          \\
Llama-2-7B  & $4$             & $4$        & $2\times 10^{-5}$  \\
\midrule
\multicolumn{4}{l}{\textbf{\textit{rule-following   fine-tuning}}} \\
GPT-3.5     & $4$             & $4$        & OpenAI API default value \\
Llama-2-7B  & $1$             & $8$        & $2\times 10^{-5}$  \\
\bottomrule
\end{tabular}
\caption{Hyper-parameters}
\label{tab:hyperparams}
\end{table}

\section{In-context Learning}
\label{sec:ICL}
\subsection{Case-based Reasoning in ICL}
We have discussed two reasoning mechanism of case-based reasoning and rule-based reasoning in experiments of fine-tuning LLMs. However, another crucial aspect of LLMs' reasoning ability is attributed to in-context learning (ICL). This method draws upon knowledge not only ingrained during the pre-training but also from specific examples supplied within the context. The underlying mechanisms that make ICL effective are among the most intriguing and unanswered questions in the field. In this section, we extend our investigations to ICL, revealing that LLMs' ICL reasoning ability also exhibits characteristics of case-based learning.

Because ICL is an emergent ability~\citep{gpt3}, we choose a stronger model: GPT-3.5-turbo-0125. We use the \textit{base addition} task where we randomly add two base-9 integers with 3 digits. The adopted GPT-3.5 can rarely solve the task only with a task description, making sure that the investigated reasoning power comes from ICL. See the zero-shot results in Appendix~\ref{app:in_context_zero_shot}.

To study whether ICL reasoning relies on rules or similar cases in the context, we randomly collected pairs of base-9 integers whose zero-shot addition accuracy is less than $20\%$.
Then, we provide 10 few-shot examples with the correct answers for each pair of integers, five of which are \textit{randomly} selected (called random group), and another five are obtained by simultaneously replacing only one digit of the pair (thus are considered as more similar examples than the first five, and called similar group). Scratchpad is used in each few-shot example to provide step-by-step intermediate results.
We choose the 14 test samples where the improvement with few-shot examples is more than $80\%$. 

\begin{figure*}
    \centering
    \includegraphics[width=1\linewidth]{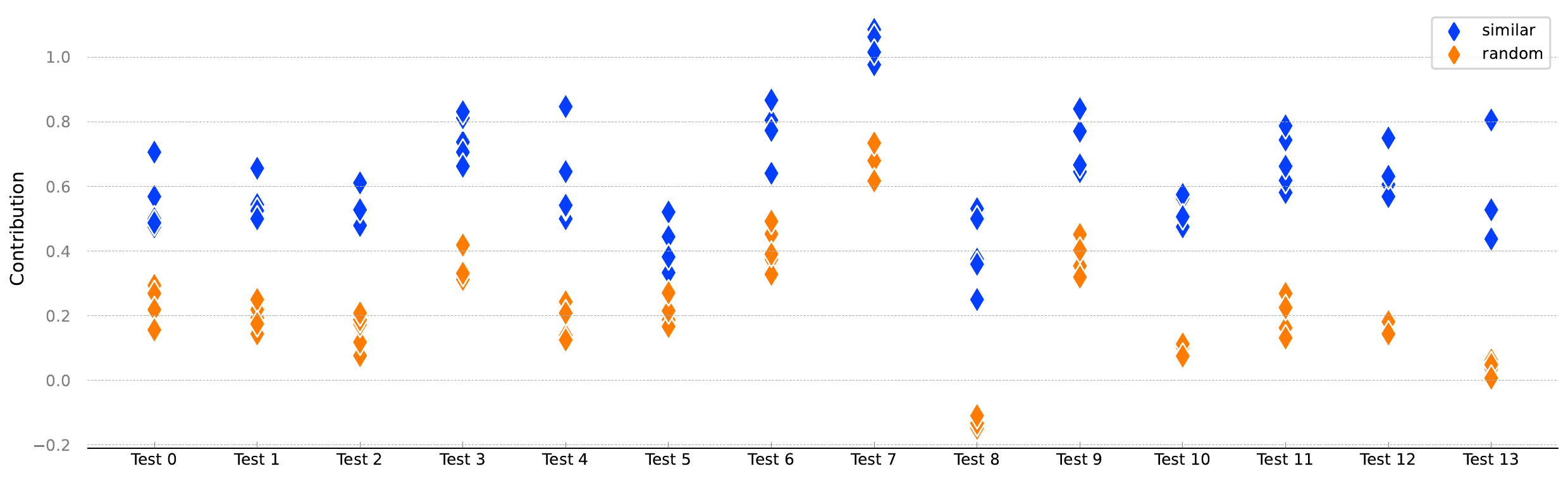}
    \caption{The contribution of \textit{similar} and \textit{random} ICL examples in 14 tests. For contribution in each experiment, the contribution of similar examples is significantly larger than that of random ones.}
    \label{fig:icl_complete}
\end{figure*}

To determine the contribution of each example, we adopt an intervention experiment similar to \S\ref{method} where we mask some in-context examples from either the similar group or the random group and compare the accuracy drop.
Considering the interaction between individual examples, we choose to traverse all mask possibilities within a group instead of masking only one example. For example, for the similar group, we will have $2^5-1=31$ possible masks (excluding the empty mask).
Specifically, we measure the contribution of the $i$-th in-context example as 
\begin{equation*}
    c_i = \left.\left(\sum_{m\in\mathcal{M}}\mathds{1}\{i\in m\}\cdot \frac{accu_{m}-accu_{orig}}{accu_{icl}-accu_{orig}}\right) \middle/ N_i\right.,
\end{equation*}
where $accu_{orig}$, $accu_{icl}$ and $accu_{m}$ represent the accuracy without few-shot examples, with all examples and with non-masked examples, respectively. $\mathcal{M}$ is the mask set that contains all possible combinations in the random and similar group. $N_i$ is the number of masks that contain $i$ (which is a constant 16).
We report the contribution of similar and random examples to the 14 test samples in Figure~\ref{fig:icl_complete}. 
The contribution of the similar group is significantly greater than that of the random group in all experiments, with the p-value of 14 average values $<0.001$.


These results suggest that the model relies more on directly discovering shortcuts from similar cases rather than summarizing the reasoning rules of the task.
This phenomenon seems contrary to some previous views on ICL which point out that the contribution of in-context examples lies mainly on hints about ``tasks'' and ``domains'' rather than specific functions, implying a more rule-based method. We believe the difference comes from the basic capacity of LLMs to solve the task. For some tasks where LLMs have captured the essential reasoning abilities, ICL examples may help them ``recall'' the task so that the model can benefit from even some dissimilar examples. In contrast, when the model is unfamiliar with the task, it is difficult to solve the problem through recalling the pre-training knowledge. In this case, only similar examples can improve model performance by providing more direct shortcuts.
In a word, our experiments suggest that it may not be possible to expect the model to extract rules that were not obtained during the pre-training phase by summarizing ICL examples.

\subsection{Zero-shot Results}
\label{app:in_context_zero_shot}
Base addition is an ``unfamiliar'' task that the model cannot solve without few shot examples. To show it, we test GPT-3.5-turbo-0125 with 100 pairs of base 9 integers with 3 digits. We use the system message as ``You are a helpful assistant to solve arithmetic problems. You will be provided with two base 9 integers and you need to return the sum of the two integers in base 9.'' the user message as `` Int a: \textit{a}; Int b: \textit{b}.'' The model can happen to generate the correct answer by summing two numbers in base 10. For example, 236+321, which is equal to 557 in either base 10 or base 9. So we also test the model on the test set where these easy samples are removed. The results is shown in Table~\ref{tab:in_context_zero_shot}.

\begin{table}[H]
\centering
\begin{tabular}{lcc}
\toprule
Task     & 0-shot base addition & 0-shot base addition (hard) \\ \midrule
Accuracy &  $8.8\%\pm 10.5\%$    & $8.0\%\pm 10.0\%$             \\ \bottomrule
\end{tabular}
\caption{Zero-shot accuracy of GPT-3.5-turbo on base addition}
\label{tab:in_context_zero_shot}
\end{table}


\subsection{Rule Following Ability from In-context learning}
\label{app:in_context_learning_rf} 
\subsubsection{Addition}

\begin{figure}[H]
    \centering
    \includegraphics[width=.4\textwidth]{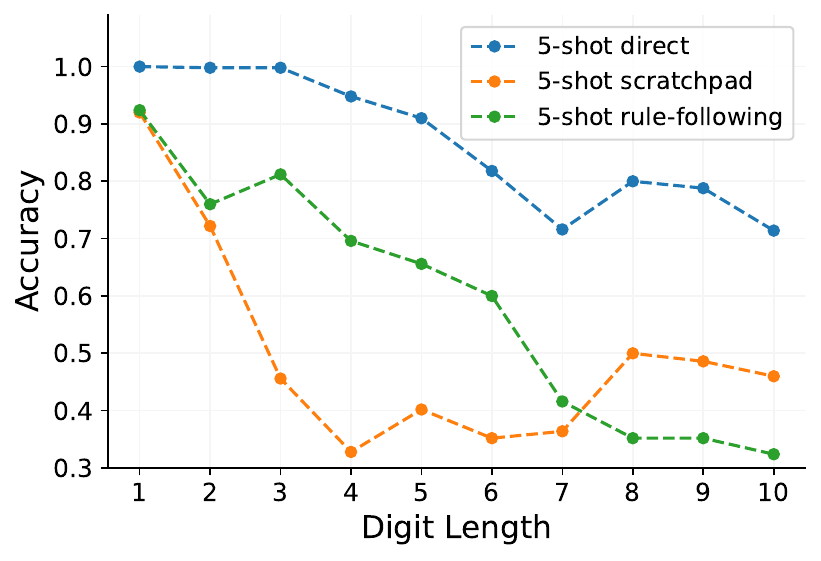}
    \includegraphics[width=.4\textwidth]{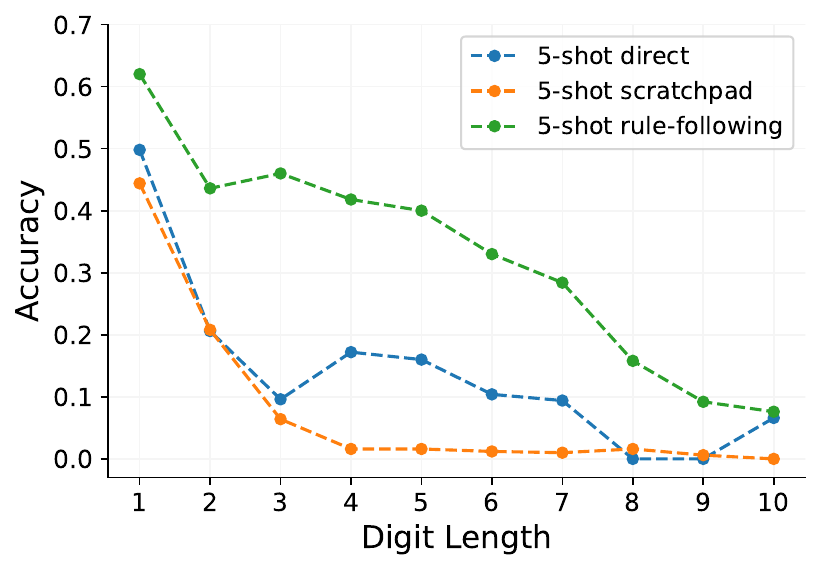}
    \caption{In-context learning performance on \textbf{addition} task (\textbf{left}) and \textbf{base addition} task (\textbf{right}).}
    \label{fig:in-context}
\end{figure}

We first conduct experiments on a the standard base-10 addition, which is familiar to GPT-3.5. Utilizing the GPT-3.5-turbo-0125 model, for the maximal digit length among these two integers from 1 to 10, we randomly selected 100 pairs of integers  respectively. Each test pairs repeat 5 computations to obtain the average accuracy. We provided identical in-context examples for all inputs, consisting of 5 examples with a maximum length of 5 digits. These in-context examples are presented in \textit{direct}, \textit{scratchpad}, and \textit{rule-following} formats. The accuracy of each digit under these three formats is illustrated in Figure~\ref{fig:in-context} left. Rule-following prompting lags behind directly asking the model to generate the answer. This may be because the model may find shortcuts to do addition as it has been trained on a huge corpus containing various addition calculations.

\subsubsection{Base Addition}
Then, we test the performance on the base-9 addition task. Here we provides 5 examples with maximal digits 5 in direct answer, scratchpad and rule-following format, shown in Figure~\ref{fig:in-context} right. On this task, rule-following still shows good performance, but the performance of direct and scratchpad is greatly reduced, compared to addition task. This shows that in both direct and scratchpad prompt modes, the model still relies heavily on its basic capabilities. Therefore, the rule following method is particularly suitable for complex and unfamiliar tasks. This kind of detailed and clear rule guidance helps the model quickly master a certain degree of reasoning with little knowledge of the corresponding task, but for tasks where the model has learned some shortcuts, it may not help performance. The shortcut learned in the pre-training stage cannot help each other with the rules in ICL. Therefore, on the latter task, if you want the model to follow the rules for reasoning, finetuning is necessary.

\section{Additional Results of Leave-Square-Out}
\label{sec:additional_results}
\begin{figure}[H]
  \centering
  \subfigure[1 hole GPT-2]{
    \label{fig:1holes_candr_gpt2} 
    \includegraphics[width=.23\textwidth]{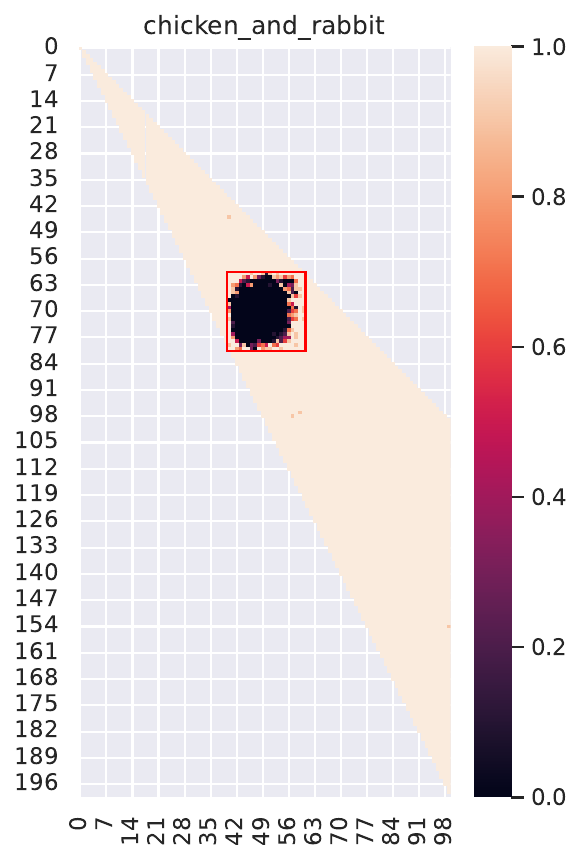}}
  \subfigure[1 hole GPT-2 Medium]{
    \label{fig:1holes_candr_gpt2-medium} 
    \includegraphics[width=.23\textwidth]{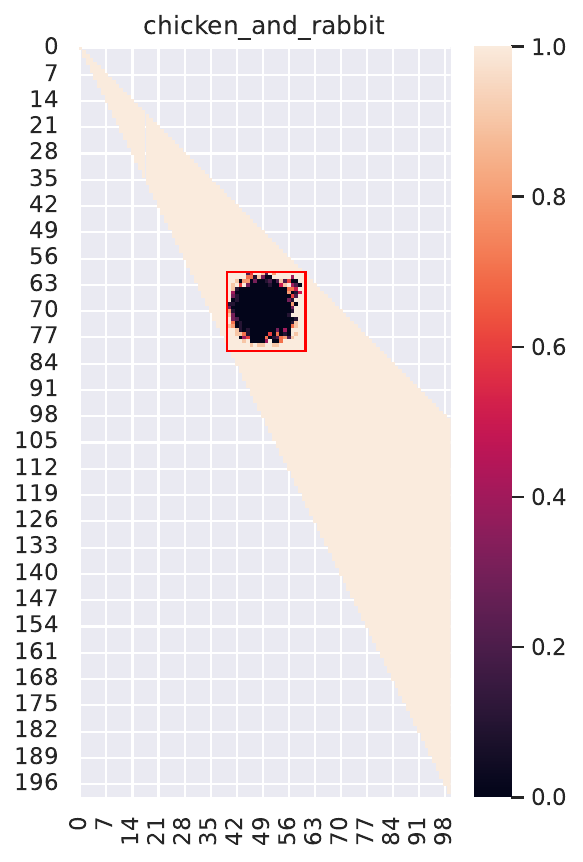}}
  \subfigure[3 holes GPT-2]{
    \label{fig:3holes_candr_gpt2} 
    \includegraphics[width=.23\textwidth]{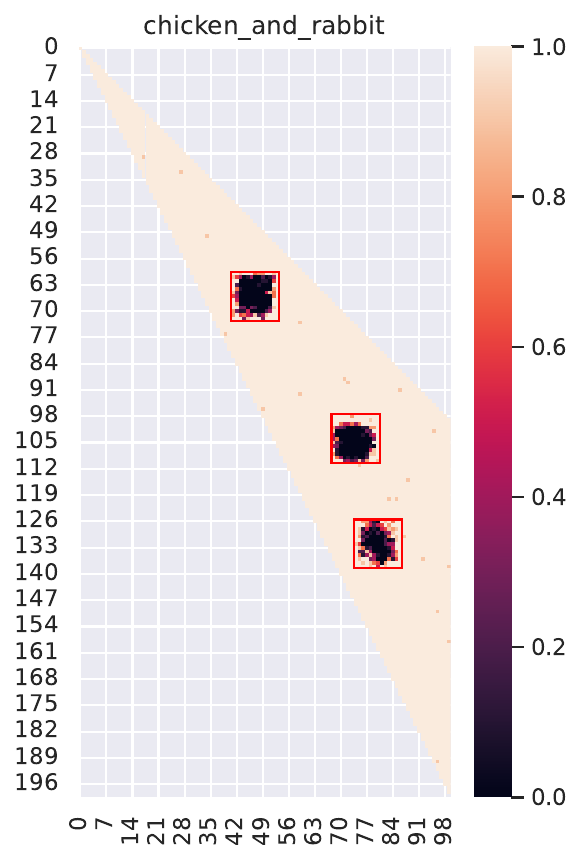}}
  \subfigure[3holes GPT-2 Medium]{
    \label{fig:3holes_candr_gpt2-medium} 
    \includegraphics[width=.23\textwidth]{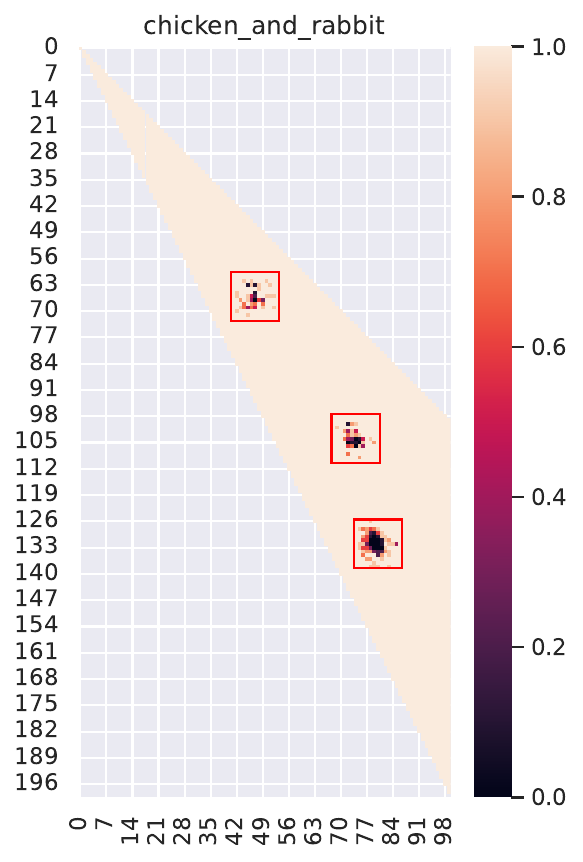}}
  
  \caption{Accuracy distributions of GPT-2 and GPT-2 Medium on chicken \& rabbit problem.}
  \label{fig:chicken_and_rabbit} 
\end{figure}

\paragraph{Chicken \& rabbit problem} We show the results of leaving test squares out of the datasets chicken and rabbit problem in Figure~\ref{fig:chicken_and_rabbit}. We experiment on two models including GPT-2 and GPT-2 Medium. The center and the length of the test square in the experiment of leaving 1 square out is $(70, 50)$, $l_k=20$. The lengths of the test squares in the experiment of 3 holes are randomly sampled in $[10, 30)$.

\begin{figure}[H]
    \centering
    \includegraphics[width=.9\textwidth]{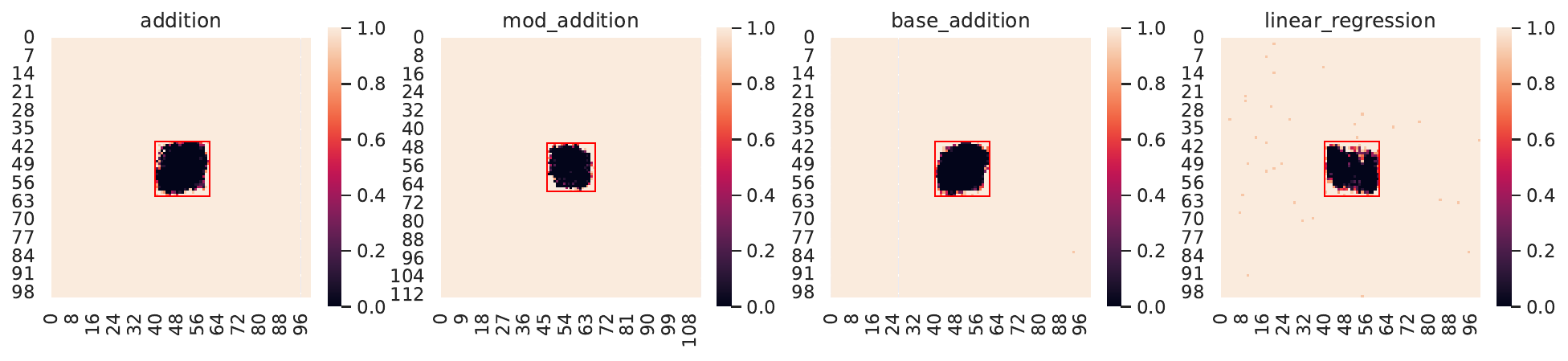}
    \caption{Accuracy of leaving a test square of length $l_k=20$ out on GPT-2 Medium.}
    \label{fig:1hole_gpt2-medium}
\end{figure}

\begin{figure}[H]
    \centering
    \includegraphics[width=.9\textwidth]{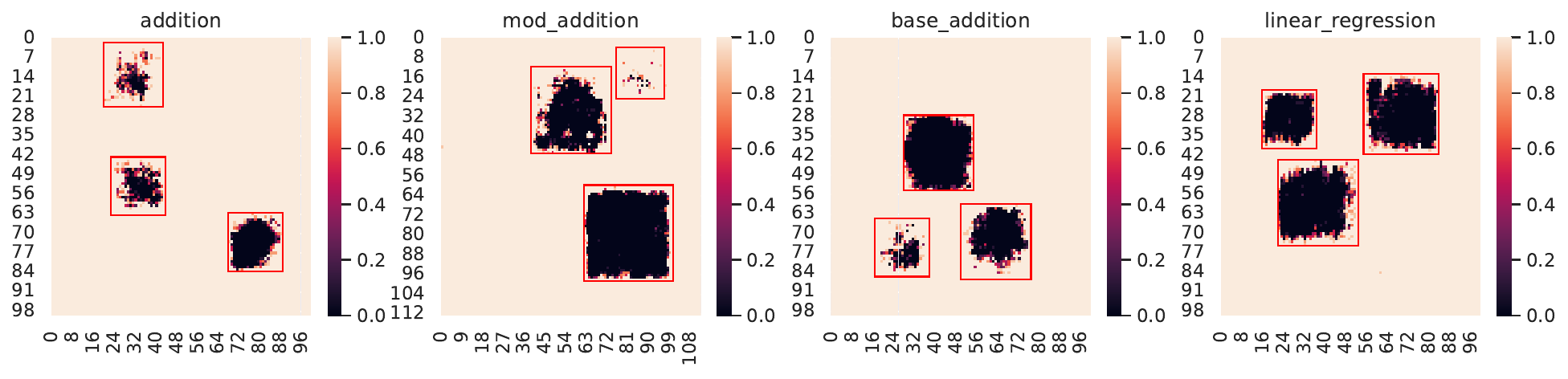}
    \caption{Accuracy of leaving 3 test squares out on GPT-2 Medium.}
    \label{fig:3holes_gpt2-medium}
\end{figure}

\begin{figure}[H]
    \centering
    \includegraphics[width=.9\textwidth]{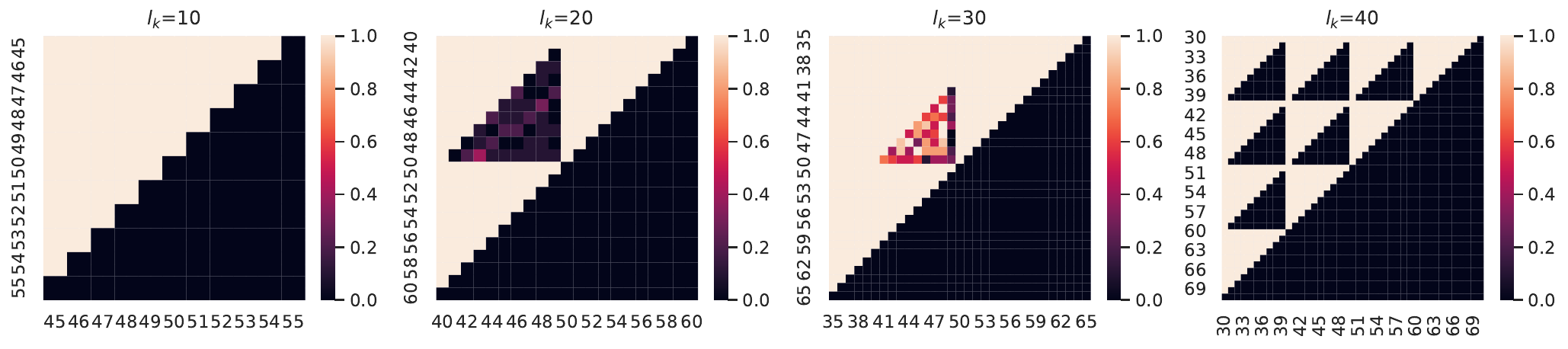}
    \caption{Test accuracy of GPT-2 Medium trained on input-output pairs containing scratchpads on the task of addition.}
    \label{fig:scratch_gpt2-medium}
\end{figure}

\paragraph{GPT-2 medium} We show the results of performing Leave-Square-Out on GPT-2 Medium on datasets including addition, modular addition, base addition and linear regression in Figure~\ref{fig:1hole_gpt2-medium} (leaving 1 square out) and Figure~\ref{fig:3holes_gpt2-medium} (leaving 3 square out). Besides, we show the results of leaving a square out on GPT-2 Medium trained on input-output pairs containing scratchpads in Figure~\ref{fig:scratch_gpt2-medium}.

\paragraph{Training curve}

\begin{figure}[]
  \centering
  \subfigure[Training loss of GPT-2.]{
    \label{fig:gpt2-loss} 
    \includegraphics[width=.95\textwidth]{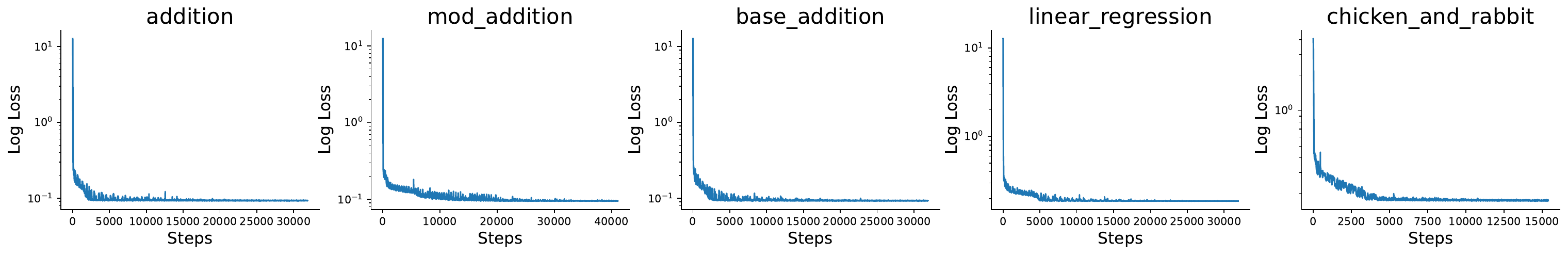}}\\
  \subfigure[Training loss of GPT-2 Medium.]{
    \label{fig:gpt2-medium-loss} 
    \includegraphics[width=.95\textwidth]{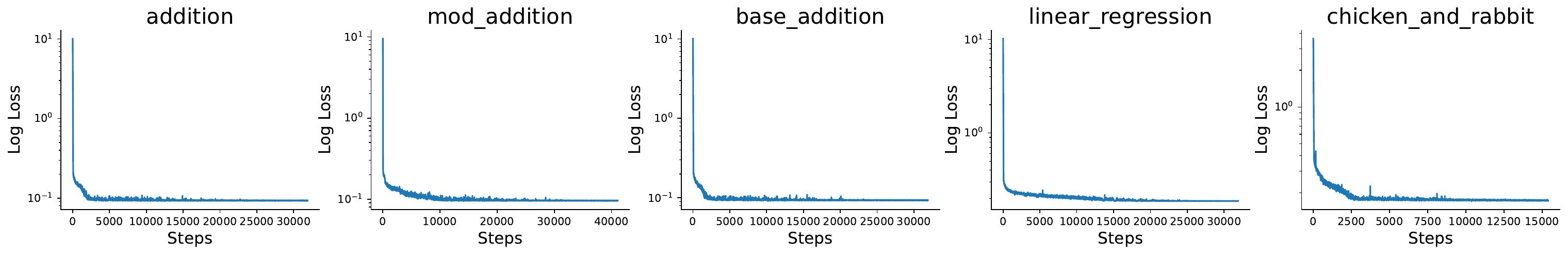}}
\caption{Log training loss of GPT-2 and GPT-2 medium in the Leave-Square-Out experiments.}
\label{training_curve}
\end{figure}

We show the training loss of GPT-2 and GPT-2 medium in the Leave-Square-Out experiments in Figure~\ref{training_curve}. 

\begin{figure}
    \centering
    \includegraphics[width=.95\textwidth]{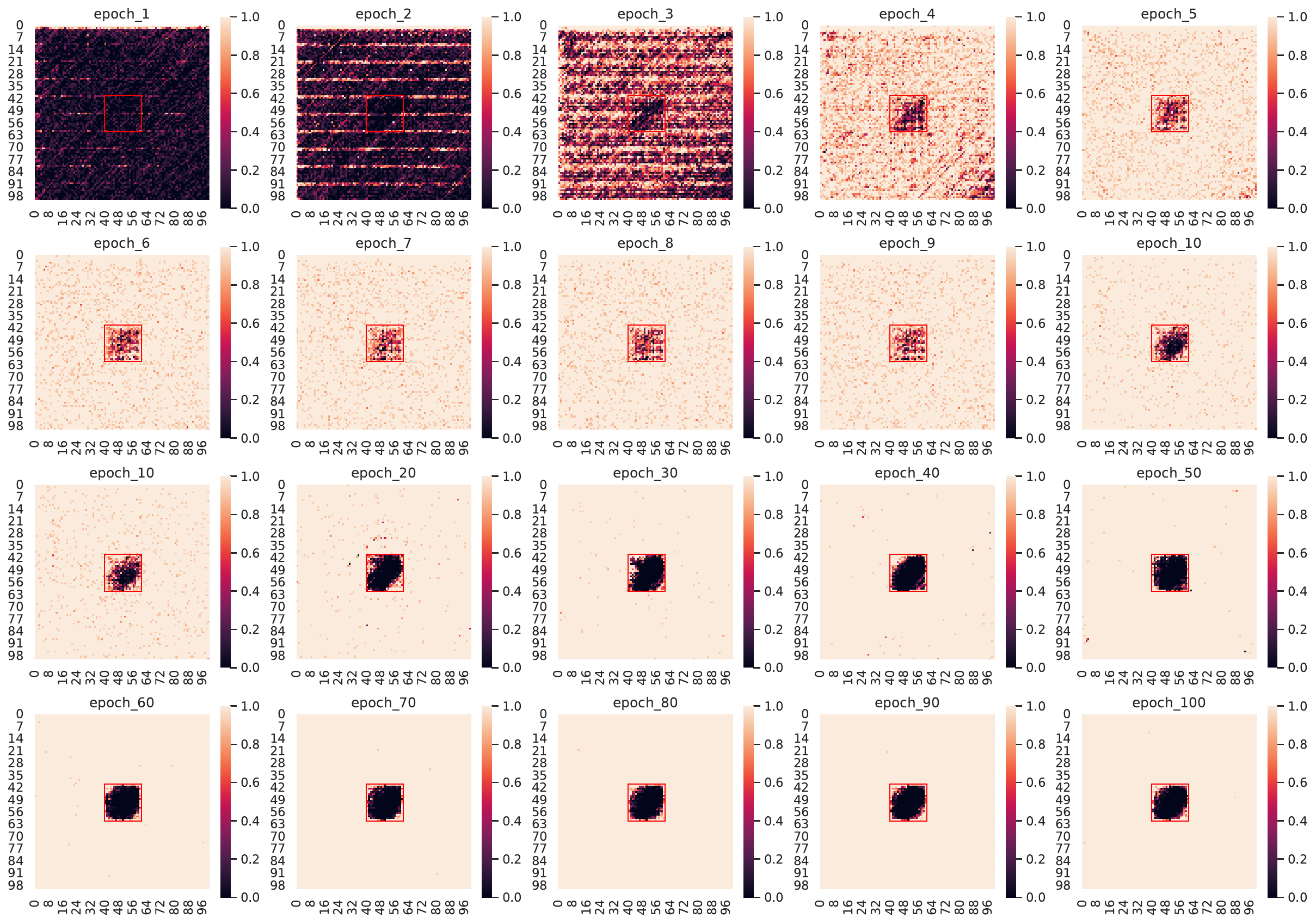}
    \caption{Model performance of GPT-2 fine-tuned after different num of epochs on the whole addition dataset. The area inside the red box represents the test square.}
    \label{fig:epoch}
\end{figure}

Besides, to have a clearer look into the training process, we conduct Leave-Square-Out experiments by fine-tuning GPT-2 on the addition task for 1, 2, 3,..., 10, 20, 30,..., 100 epochs, respectively. The results are in Figure~\ref{fig:epoch}. The center of the test square $(a_k, b_k)$ is set to $(50, 50)$, and the length $l_k$ is 20. During generation, we set the model temperature to 1 and sample 10 generations to evaluate the accuracy on each test point.

The results show that after the model's training loss is lower than a certain value (after epoch 4), the model exhibits obvious case-based reasoning behavior (holes appear in the test square). In the earlier epochs like epoch 1 and 2, the training has not saturated, thus both training and test accuracy are extremely low, which also indicates the necessity of fine-tuning for such tasks.

\begin{figure}[H]
    \centering
    \includegraphics[width=.35\textwidth]{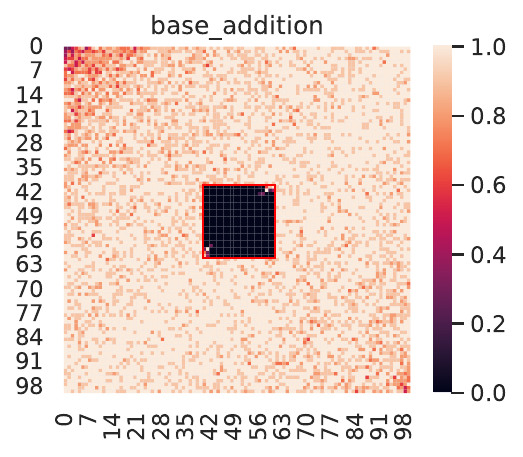}
    \caption{Accuracy of GPT-2 trained from scratch on the task of base addition when we leave a test square out. The center and the length of the test square is $(50, 50)$, $l_k=20$}
    \label{fig:1hole_base_gpt2_from_scratch}
\end{figure}

\paragraph{GPT-2 trained from scratch} We additionally experiment on GPT-2 trained from scratch on the task of base addition to see how the process of pre-training affects the reasoning mechanism. We show the results of leaving one test square with length $l_k=20$ and center $(50, 50)$ out in Figure~\ref{fig:1hole_base_gpt2_from_scratch}. Besides, we conduct the experiments of training the model in the random-split setting with training set accounting for 70\% of the whole dataset. The model can achieve more than 98\% accuracy on the test set in the random-split setting. As shown in Figure~\ref{fig:1hole_base_gpt2_from_scratch}, there is a black hole in the test square, suggesting the behavior of case-based reasoning is still obvious when we directly train the model from scratch. Besides, we observe in the training process that the training loss converges much more slowly than in the setting of fine-tuning a pre-trained model.

\begin{figure}
    \centering
    \includegraphics[width=.6\textwidth]{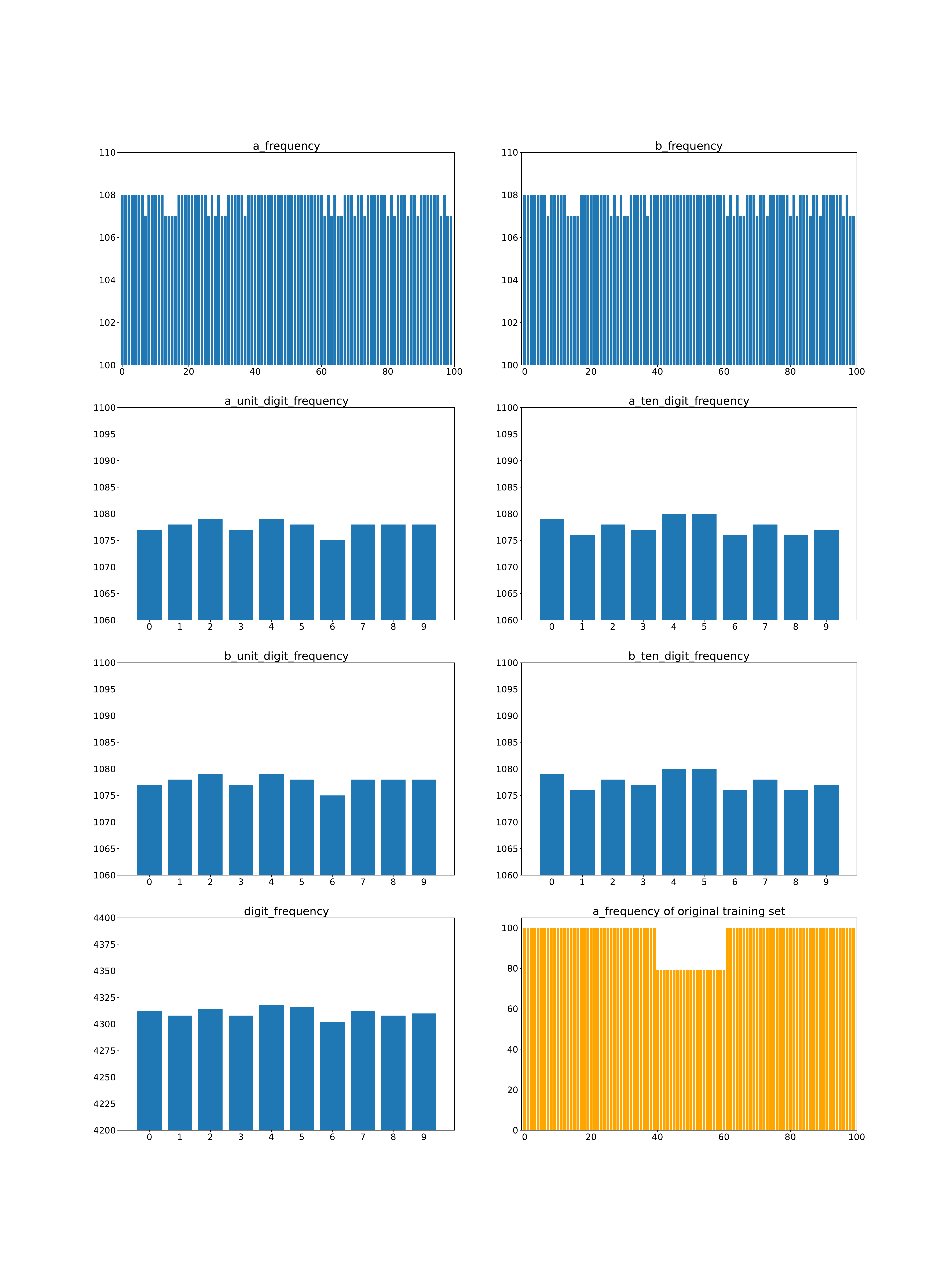}
    \caption{The frequency of numbers $a$ and $b$, the frequency of unit digits and tens digits of $a$ and $b$, and the frequency of digits in both $a$ and $b$. We show the digit frequency after upsampling with blue histograms and that before upsampling with orange histograms.}
    \label{fig:digit_frequency}
\end{figure}

\paragraph{Considerations for class imbalance}
As some may worry that the method of leaving a test square out may cause data imbalance, thus confusing the model, we conduct an additional experiment to study the effect of class imbalance issue by upsampling those numbers that originally occur less in the training set of the addition task, thereby maintaining a balanced distribution of numbers and digits in the new training set. In Figure~\ref{fig:digit_frequency}, we show the digit frequency before and after upsampling. It shows that both the numbers and digits are balanced after upsampling.

\begin{figure}
    \centering
    \includegraphics[width=.35\textwidth]{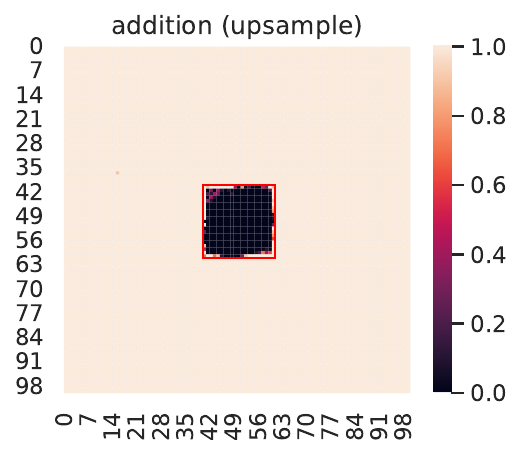}
    \caption{Performance of fine-tuned GPT-2 on addition after upsampling.}
    \label{fig:gpt_2_upsample}
\end{figure}

Then, we fine-tune the model on this updated training set and repeat the experiment of Figure~\ref{fig:holes}. The new results are illustrated in Figure~\ref{fig:gpt_2_upsample}. As we can see, the hole still appears, demonstrating the behavior of case-based reasoning. This indicates that class imbalance is not a confounder of our results.

Besides, we also show the frequency of number a in the original training set in Figure~\ref{fig:digit_frequency}. It should be noticed that our original training set is not a dataset with extreme data imbalance, as we only leave 20 out of 100 samples of certain numbers.

\section{Ablations for Leave-Square-Out}
\subsection{Ablation for Data Size}
\label{app:data_size}
\begin{figure}[H]
    \centering
    \includegraphics[width=.9\textwidth]{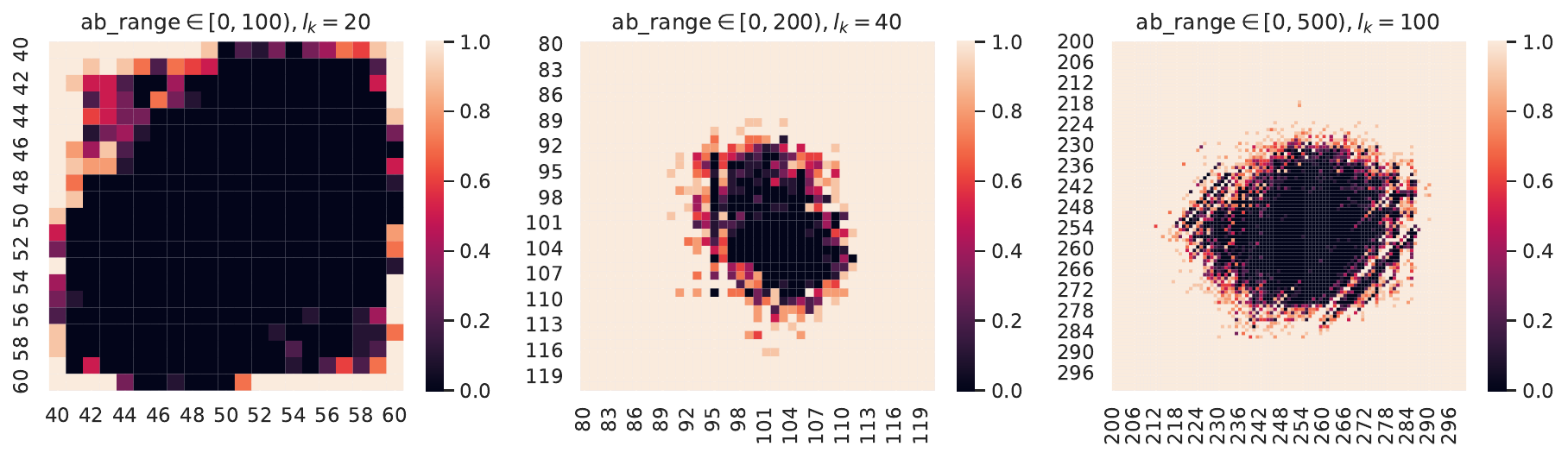}
    \caption{Test accuracy of models trained on addition of different data size.}
    \label{fig:data_size}
\end{figure}

To explore the effects of datasize on the model behavior, we conduct experiments of Leave-Square-Out on the task of addition of different range of $a, b$, including $[0,100)$, $[0,200)$, $[0,500)$. Correspondingly, we scale up the length of test square to be $l_k=20, 40, 100$ respectively. We use GPT-2 and use the same hyper-parameters in each dataset. We train the model with 100 epochs, batch size set to 30 and learning rate set to $10^{-4}$. The results are shown in Figure~\ref{fig:data_size}. Holes can still be observed in the setting where $a, b\in[0, 500)$ (the dataset scales up 25 times), suggesting that the models are still doing case-based reasoning when the data size scales up.

\subsection{Ablation for Model Size: Llama-2-7B}
\label{app:model_size_llama}
To show the effects of model size on case-based reasoning, we conduct experiments on Llama-2-7B on the task of base addition because Llama-2 has already learned addition to a some degree. To further eliminate the effect of pre-training, we train the model \textit{from scratch} instead of fine-tuning. 

\begin{figure}[H]
  \centering
  \subfigure[Accuracy in the \textbf{test square} of Llama-2-7B on the task of base addition. Center $(50, 50)$, length $l_k=20$.]{
    \label{fig:1hole_base_llama2_from_scratch} 
    \includegraphics[width=.25\textwidth]{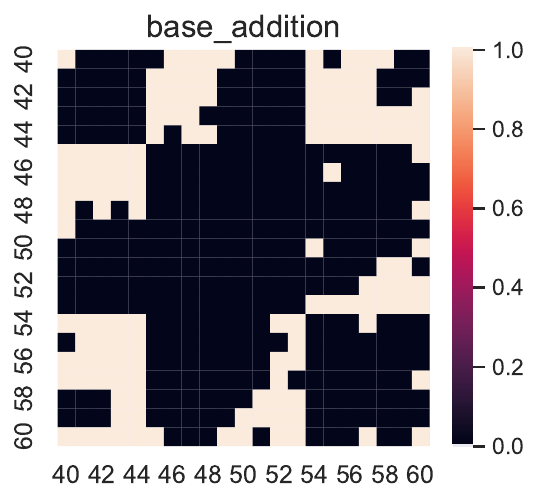}}
    \quad  \quad
  \subfigure[Test accuracy distribution of Llama-2-7B on the task of base addition over $a$ and $b$ when we leave a test square of side length $l_k=140$ and center $(350, 350)$.]{
    \label{fig:llama_base_ood_70} 
    \includegraphics[width=.25\textwidth]{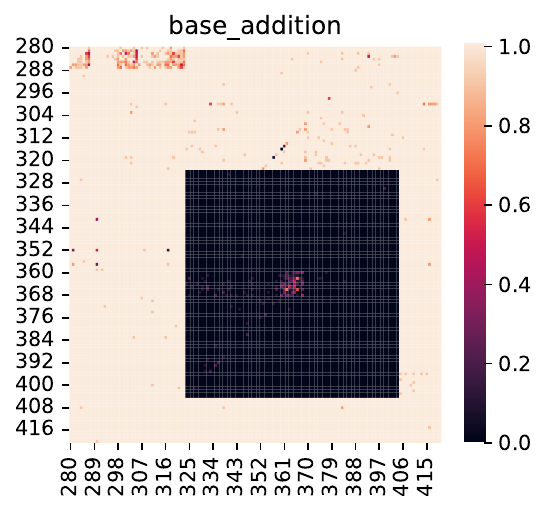}} \quad  \quad
  \subfigure[Accuracy of Llama-2 trained from scratch on the task of base addition when we leave a test square out. The center and the length of the test square is $(428, 428)$ (represented in base-9), $l_k=140$]{
    \label{fig:llama_base_ood_70_base9} 
    \includegraphics[width=.25\textwidth]{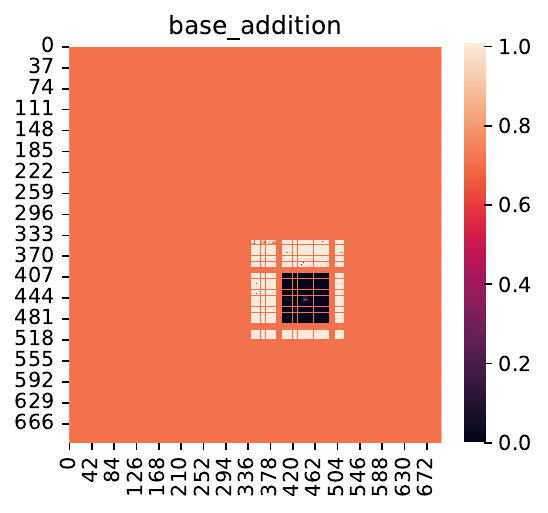}}
  \caption{Accuracy of Llama-2-7B \textbf{trained} on base addition \textbf{from scratch}.}
\end{figure}

We maintain the same data size as GPT-2, i.e., setting the range of $a$ and $b$ to $[0,100)$ and leave out a test square with center $(50, 50)$ and side length $l_k=20$. 
Despite this, the test accuracy in the Leave-Square-Out setting is 30.2\%, far lower than the random split accuracy. As the comparison, the test accuracy in the random split setting reaches $92\%$. There are still holes in the accuracy distribution of the test squares, as shown in Figure~\ref{fig:1hole_base_llama2_from_scratch}. 

Note that different from GPT-2, the test accuracy in the random split setting cannot reach 100\% even after training for 500 epochs where training loss has almost converged, suggesting overfitting.
In light of this, we also conduct an experiment where we correspondingly enlarge the range of $a$ and $b$ to $[0, 700)$ with the side length of test square $l_k=140$ and center $(350, 350)$, forming a training set accounting for about 96\% of the whole dataset. We first conducted the experiment in the random-split setting with 70\%-30\% training-test ratio and verified the model can reach 100\% accuracy. Then we perform the Leave-Square-Out experiment. Figure~\ref{fig:llama_base_ood_70} shows the results. The model still demonstrates significant case-based reasoning behavior by failing to answer a large portion of test samples. This indicates that the trend of case-based reasoning still exists when the model scales up. Furthermore, we also plot it with base-9 coordinates in Figure~\ref{fig:llama_base_ood_70_base9}, which shows a highly structured pattern possibly related to the task structure. Here, we provide a preliminary analysis in Appendix~\ref{app:error_analysis}, leaving a more in-depth exploration for future work.

\begin{figure}[H]
  \centering
  \subfigure[Accuracy of test square in the Leave-Square-Out experiment.]{
    \label{fig:finetuned_llama_on_base_addition_lso} 
    \includegraphics[width=.35\textwidth]{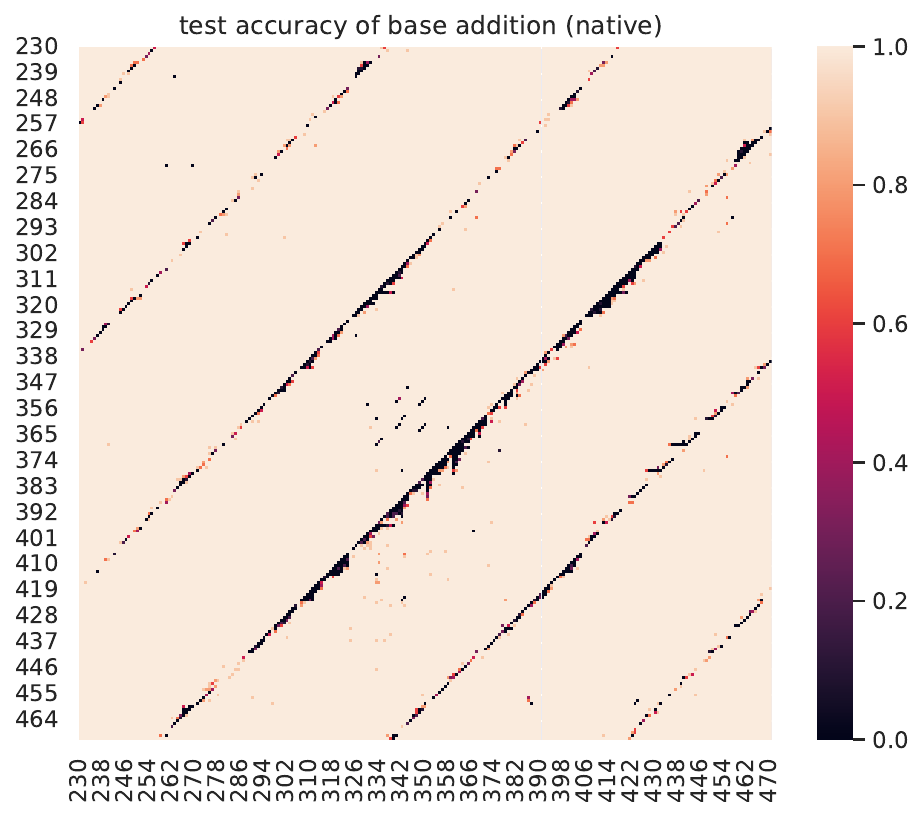}}
    \quad  \quad
  \subfigure[Accuracy of the whole dataset in the random split experiment.]{
    \label{fig:finetuned_llama_on_base_addition_random} 
    \includegraphics[width=.35\textwidth]{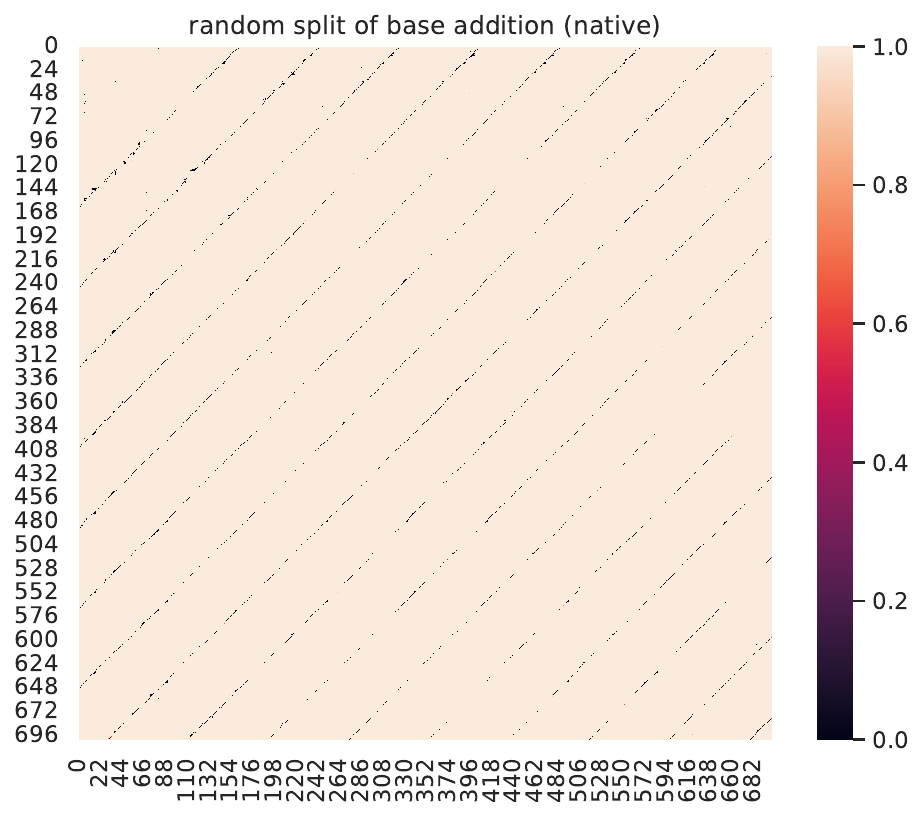}}
\caption{Performance of Llama-2-7B \textbf{fine-tuned} on base-9 addition.}
\label{fig:finetuned_llama_on_base_addition}
\end{figure}

\paragraph{Fine-tuning or train from scratch?}
Also, we conduct experiments of fine-tuning pretrained Llama-2-7B on base-9 addition. The results are shown in Figure~\ref{fig:finetuned_llama_on_base_addition}.

Firstly, we observe that the results are indeed different from training from scratch. There are no obvious holes in the test square. Instead, model performance drops in areas of anti-diagonals of both training and test regions.

Admittedly, the results do not provide evidence for case-based reasoning, however, it \textbf{does not necessarily indicate that the model is performing rule-based reasoning} either. The reasons are as follows:

\begin{enumerate}
    \item It is possible that the left-out square are \textbf{not really the dependent cases} for pretrained Llama-2-7B on this task. Our hypothesis that surrounding cases are the dependent cases may not hold for this setting. It is possible that other training/test spliting method can reveal case-based reasoning behavior again.
    \item There might be \textbf{data leakage} during pretraining. Without considering data leakage, pretraining may still have introduced strong biases that happen to suit base addition well, making the model generalize to most parts of the test square. It should be noticed that introducing biases is essentially different from enhancing the model's foundamental reasoning abilities or equipping it with the ability to perform rule-based reasoning after fine-tuning, because the biases may only suit some specific tasks or representations, rather than uniformly helping models to learn rules for different tasks/representations (as will be discussed in the following base-9 addition with exotic digits experiment).
    \item We test the fine-tuned model on OOD samples involving 4/5/6 digits. The results are listed in Table~\ref{tab:performance_of_llama-7b}. The results indicate that the model at least \textbf{does not learn rules that can generalize across different lengths}. In other words, the model might learn some shortcuts working on same-length samples, but fail to learn the most faithful base addition rules that allows for length generalization.
\end{enumerate}

\begin{table}
    \centering
    \begin{tabular}{c|ccc}
    \toprule
        Digit Length & 4-digit & 5-digit & 6-digit \\
    \midrule
        Accuracy & 91.88\% & 74.14\% & 51.46\%\\
    \bottomrule
    \end{tabular}
    \caption{Performance of Fine-tuned Llama-2-7b on OOD samples of base addition. We test 5,000 samples.}
    \label{tab:performance_of_llama-7b}
\end{table}

To further investigate the reasoning mechanism of pretrained Llama-2-7B and mitigate the risk of data leakage, we alter the representations of base-9 numbers and fine-tune Llama-2-7B on the new and more challenging task. Specifically, we \textbf{replace the digits 0-8 with letters A-I} to create a counterfactual dataset which has little chance to have been exposed to the pretrained model. For example, we replace ``\texttt{1+8=10}'' with ``\texttt{B+I=BA}''. We call the new task with replaced digits base-9 addition with ``exotic digits''.

\begin{figure}
    \centering
    \includegraphics[width=.35\textwidth]{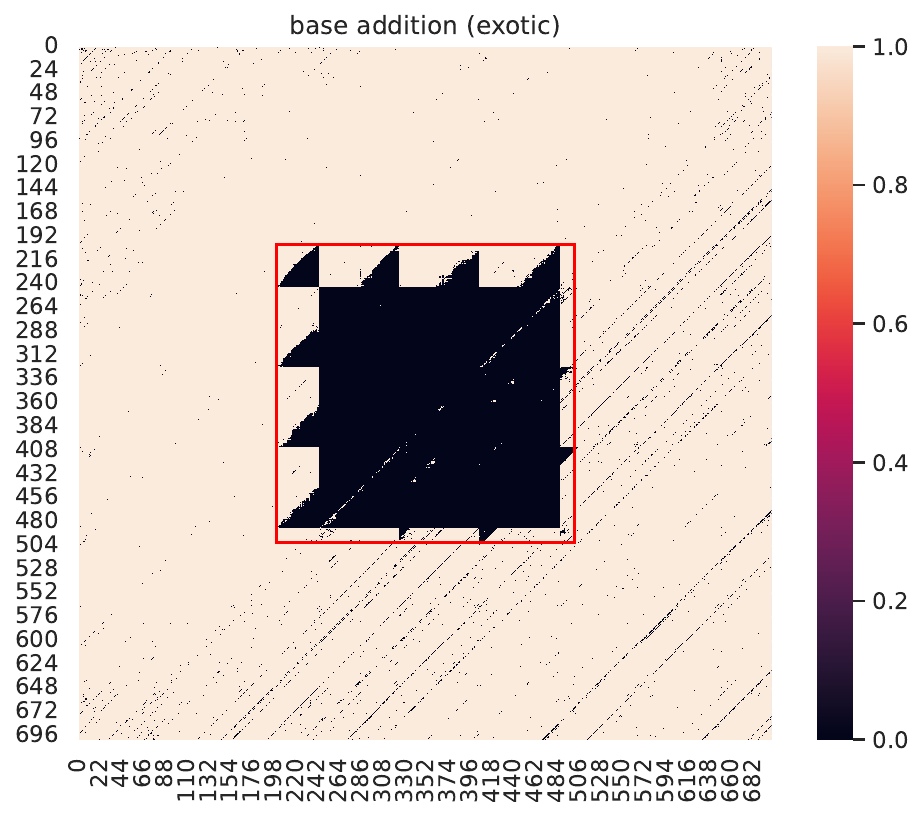}
    \caption{Performance of \textbf{fine-tuned} Llama-2-7B on base-9 addition with exotic digits. The area inside the red box represents the test square.}
    \label{fig:finetuned_llama_on_exotic_base_addition}
\end{figure}

For Llama trained from scratch, there should be nearly no distinction between these two representations if we ignore tokenization differences. For fine-tuned Llama, if it relies on a systematic mechanism to induce rules from examples, it should be able to learn these rules regardless of whether native digits or exotic digits (i.e. A-I letters) are used. However, the test performance and accuracy distribution reveal significant discrepancies from the native-digit setting, despite the training still achieves close to 100\% accuracy. We show the new accuracy distribution in Figure~\ref{fig:finetuned_llama_on_exotic_base_addition}.

As we can see, after using letter representations for base-9 numbers, there is a hole in the test square, demonstrating case-based reasoning behavior. This might indicate that the large-scale pretraining does not equip models with systematic and general rule-learning abilities. Instead, it is more likely that pretraining introduces biases suitable only for certain tasks and representations, enabling certain degree of generalization.

In conclusion, we first fine-tune pretrained Llama on regular base addition and the model performance drops in anti-diagonal areas but shows no holes, demonstrating different patterns from training from scratch. However, the phenomenon does not necessarily indicate that the model is performing rule-based reasoning. To dig deeper, we change the representations of digits into exotic letters and fine-tune Llama on the new task, which shows clear evidence of case-based reasoning. This suggests that the reasoning behavior of LLMs can be highly dependent on the input representations. Besides, large-scale pretraining seems not equip the model with the ability of systematic rule learning that can adapt to various representations.

\paragraph{Error analysis of Llama-2-7B on base addition}\label{app:error_analysis}

In the experiment of training larger model Llama-2-7B on base-9 addition task as described in Appendix~\ref{app:model_size_llama}, we find that there is still a hole in the test set, indicating that even if the models scale up, they struggle to learn to perform rule-based reasoning. Furthermore, we conduct a more detailed analysis. We observed the model usually generates wrong answers when both ``a'' and `b'' input values had hundreds of digits of ``4''. For instance, the model can correctly output ``400 + 388 = 788'', however, it failed when presented with ``400 + 400'', generating the output as ``500''. It appears to draw from the ``closeness'' between 400 and 388, failing to grasp the difference of 1 in base-9 as opposed to a difference of 12 in base-10, resulting in an erroneous output. Moreover, for the sequences ``400+401 ='', ``400+402 ='', and ``400+403 ='', the model output ``501'', ``502'', ``503'' respectively. These findings suggest that the model relies heavily on the context of closely related cases for its calculations rather than utilizing rule-based reasoning.



\subsection{Ablation for Model Size: GPT-3.5-turbo}
\label{app: model_size_gpt3.5}
To verify whether the conclusions are consistent for larger models, we conduct additional experiments on GPT-3.5-turbo. To be more specific, we choose the task of base-9 addition (less likely to appear in pre-training corpus than addition) and leave a test square with center $(a_k, b_k)=(350,350)$ and length $l_k=300$ (accounting for 18.5\% of the whole dataset) out of the whole dataset with $a\in[0, 700),b\in[0, 700)$. We fine-tune GPT-3.5 for 1 epoch with batch size set to 80.

\begin{figure}
    \centering
    \includegraphics[width=.4\textwidth]{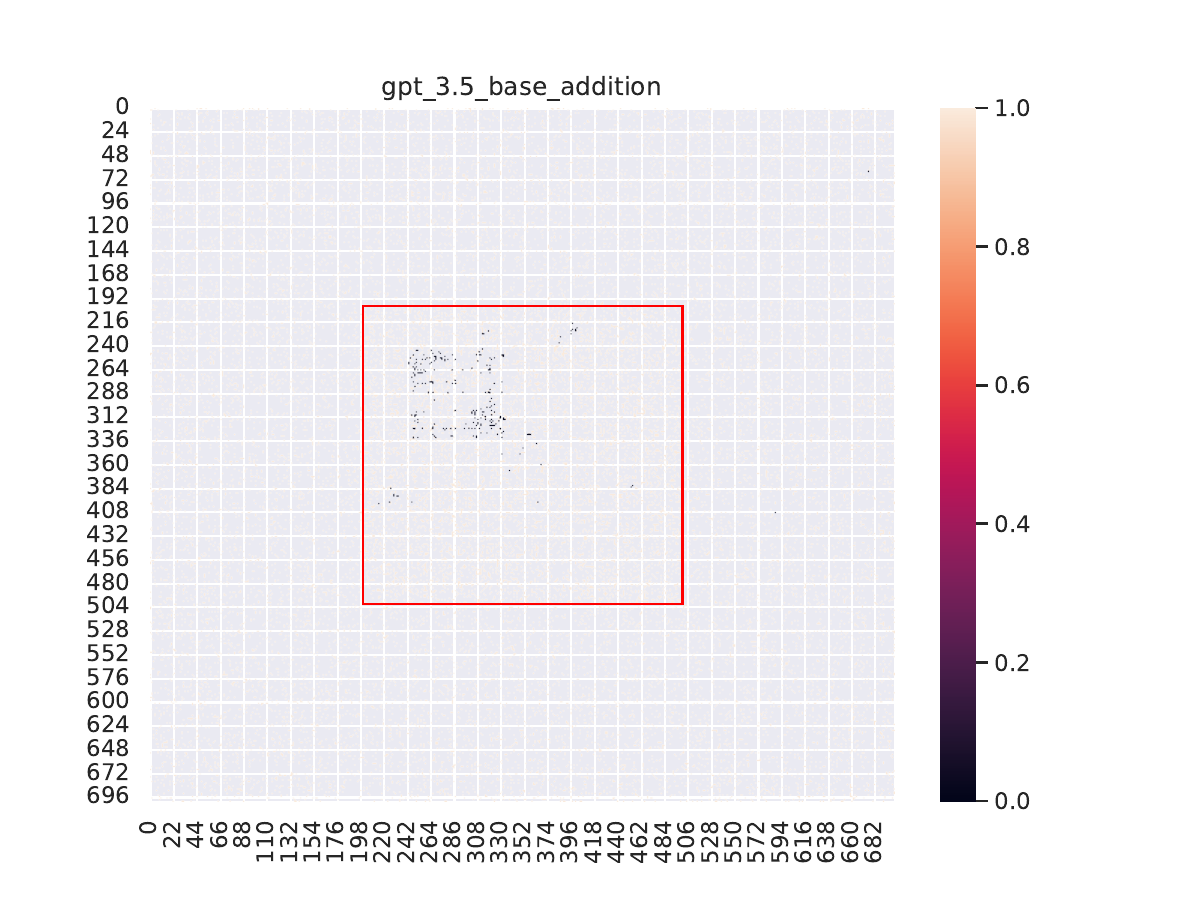}
    \caption{The training and test accuracy distribution of fine-tuned GPT-3.5-turbo on base-9 addition. The area inside the red box represents the test square. The figure needs to be scaled up to see the light pink samples that are correctly answered.}
    \label{fig:gpt_3.5_lso}
\end{figure}

The training and test accuracy distribution of the fine-tuned model is shown in Figure~\ref{fig:gpt_3.5_lso}. Due to the limited budget, we randomly sample 20\% datapoints out of the test set and 10\% datapoints out of the training set to do inference. For each sample, we perform single generation with model temperature set to 0. As we can see from the figure, there is still a ``hole'' in the test square, demonstrating case-based reasoning behavior.

More specifically, we list train/test accuracy in the Leave-Square-Out experiment and test accuracy in the corresponding random-split setting in Table~\ref{tab:gpt_3.5_acc}. In the random-split experiment, the training set accounts for 70\% of the dataset, while in the Leave-Square-Out experiment, the training set accounts for 81.5\% of the whole dataset. As can be seen in the table, although test accuracy (square) is high ($\sim$97.5\%), there are still gaps from training accuracy (square) ($\sim$100\%) and test accuracy (random) ($\sim$100\%). In other words, despite using more training data in the leave-square-out experiment, the test accuracy cannot saturate like that in the random-split experiment. In Figure~\ref{fig:gpt_3.5_lso}, we show the samples where the model generates wrong answers with black points (corresponding to the accuracy gap) to highlight the ``hole'' area where the model's performance drops. This indicates that there is still a small area where the model is relatively easy to fail, and the ability to perfectly solve the test cases in this specific area relies on training on the test square. This exactly implies case-based reasoning. Nevertheless, the hole is not as large as those in the experiments of smaller models, indicating that stronger LLMs might have better capability to leverage longer-dependency cases so that most cases in the test square may still find some dependent cases in the training set.

\begin{table}
    \centering
    \begin{tabular}{ccc}
    \toprule
        Training Accuracy (square) & Test Accuracy (square) & Test Accuracy (random) \\
        \midrule
        0.9999 & 0.9751 & 0.9998 \\
    \bottomrule
    \end{tabular}
    \caption{The left two columns shows training and test accuracy of GPT-3.5 fine-tuned in the Leave-Square-Out experiments. The rightmost column shows test accuracy of the model fine-tuned in the random-split setting.}
    \label{tab:gpt_3.5_acc}
\end{table}

\section{Training Details and Additional Results of Rule-following Finetuning}

\subsection{Training Details of Rule-following Finetuning}
\label{rule_detail}
When fine-tuning Llama, we use the same hyper-parameters for each of three different settings, excluding the model\_max\_length varied to 64, 512, and 4096 for direct, scratchpad and rule-following. To be more specific, we train the models with 1 epoch, 8 batch sizes, a dropout rate of 0.05 and a learning rate of 2e-5, saving the best model at the last step. We use greedy decoding in the inference stage to generate output with a temperature of 0.

When finetuning GPT-3.5-turbo, we use a training set of 100 samples and train the models with OpenAI API with 4 epochs, 4 batch sizes.

\begin{table}[H]
\centering
\begin{tabularx}{.5\textwidth}{lXX}
\toprule
input: & 687042+86208=  \\
\midrule
model output: & 
687042+86208,,C:0\newline
68704+8620,0,C:1 \# added 2+8+0=0\newline
6870+862,50,C:0 \# added 4+0+1=5\newline
\textcolor{red}{687+86,250,C:1 \# added 0+2+0=2}\newline
68+8,4250,C:1 \# added 7+6+1=4\newline
6+,74250,C:1 \# added 8+8+1=7\newline
+,774250,C:0 \# added 6+0+1=7\newline
774250\\
\bottomrule
\end{tabularx}
\caption{The generation of fine-tuned Llama-2-7B with scratchpad}\label{tab:fail2_scratchpad}
\end{table}

\begin{table}[H]
\centering
\begin{tabularx}{.5\textwidth}{lXX}
\toprule
input: & 496943+2382=  \\
\midrule
model output: & 
496943+2382,,C:0\newline
49694+238,6,C:0 \textcolor{red}{\# added 4+2+0=6}\newline
4969+23,26,C:1\# added 4+8+0=2\newline
496+2,326,C:1 \# added 9+3+1=3\newline
49+,9326,C:0 \# added 6+2+1=9\newline
4+,99326,C:0 \# added 9+0+0=9\newline
+,499326,C:0 \#4+0+0=4 \newline
499326\\
\bottomrule
\end{tabularx}
\caption{The generation of fine-tuned Llama-2-7B with scratchpad}\label{tab:fail1_scratchpad}
\end{table}

\begin{table}[t]
    \centering
    \begin{tabular}{c|cccc}
    \toprule
        Model & 6 digit  & 7 digit & 8 digit & 9-digit  \\
    \midrule
         Llama-2-7b &99.6 &99.4 &96.2 &90.1  \\
         Llama-2-70b &99.3 &97.8 &97.0 &89.8\\
    \bottomrule
    \end{tabular}
    \caption{Accuracy of fine-tuned Llama-2-7b and Llama-2-70b on 6-9 digit addition task}
    \label{tab:RFFT_70b_vs_7b}
\end{table}

\subsection{Ablation Study of RFFT}
\label{rfft_ablation}
We conduct ablation studies of 5 modules of RFFT, including:
1. \textbf{process}: simplify the process of each iteration (addition of each digit), outputting multiple lines of code in one step;
2. \textbf{variable}: skip recalling relevant variables before executing each line of code;
3. \textbf{rule}: remove rules from the input (but still requires reciting the used rule in each output step);
4. \textbf{caption}: remove natural language instructions such as "1. Initialize Result and Carry" from the output which correspond to comments in the code;
5. \textbf{cite}: remove line-by-line recitation of the code from the output.

It is shown in Figure~\ref{fig:ablation} that the model's performance deteriorates significantly when removing \textbf{cite} and \textbf{caption} components, especially cite. Both of the modules aid the model in recalling rules. \textbf{Variable} also has some performance impact as it helps the model reduce its reliance on distant text. On the contrary, \textbf{process} and \textbf{rule} do not have a significant impact, and in some cases, there is even a performance improvement. This may be because reducing the context length is beneficial for the model so that it can put more strength on rule execution, and an extremely detailed guidance in step-by-step rule-following (\textbf{process}) is not necessary for this task. In conclusion, \textbf{reciting the rule used in each step} and \textbf{reminding LLMs what the current step is doing} is crucial for RFFT's success, while there maybe room for simplifying the rule representation and execution, which is left for future research.

\begin{figure}
    \centering
    \includegraphics[width=.4\textwidth]{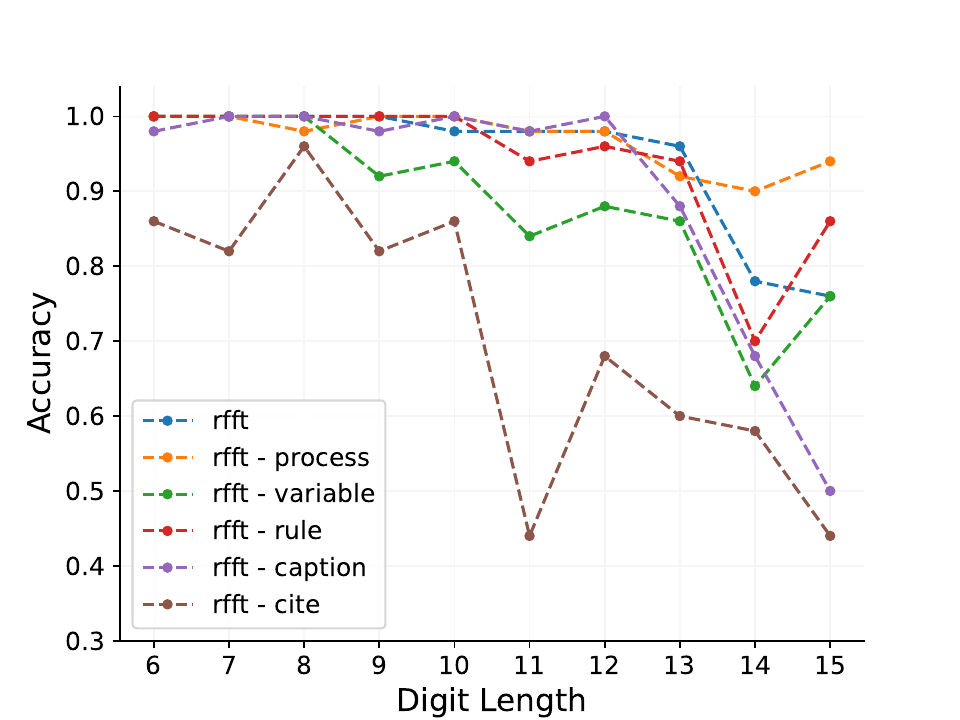}
    \caption{Ablations for RFFT.}
    \label{fig:rfft_ablation}
\end{figure}

\subsection{Rule-following as a Meta Learning Ability}\label{app:RFFT_meta}

\begin{figure}[H]
    \centering
    \includegraphics[width=.35\textwidth]{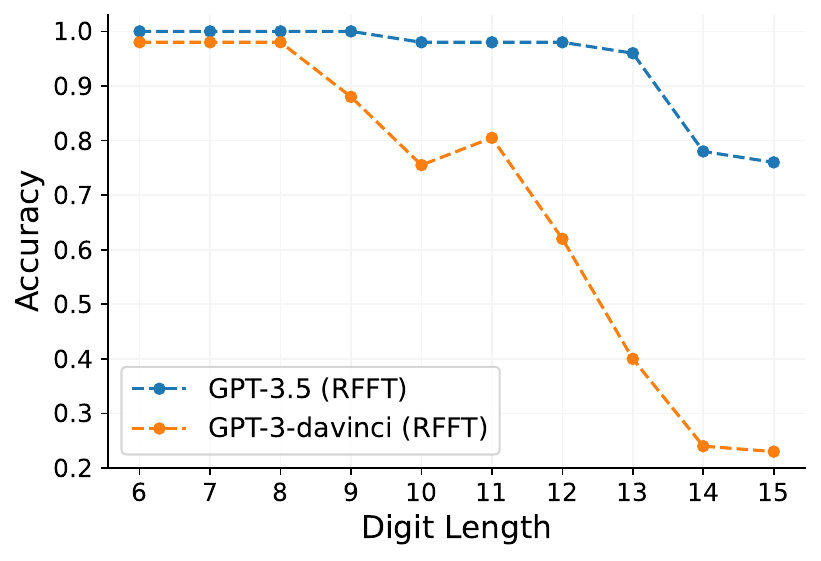}
    \caption{Accuracy of GPT-3.5 and davinci-002 fine-tuned with RFFT on addition with 6-15 digits}
    \label{fig:RFFT_3.5_vs_davinci}
\end{figure}

We fine-tune a larger model Llama-2-70b than Llama-2-7b and a slightly weaker model davinci-002 than GPT-3.5. For Llama-2-70b, we reduce the training set to 20k samples, smaller than 150k used in Llama-2-7b but achieve comparable results to Llama-2-7b as shown in Table~\ref{tab:RFFT_70b_vs_7b}. For davinci-002, we use the same data size as GPT-3.5 and Figure~\ref{fig:RFFT_3.5_vs_davinci} shows that davinci-002 can achieve up to 8 digits generalization, worse than GPT-3.5. These results indicate that models with more advanced foundational abilities can achieve better length generalization after applying RFFT, even with small data, revealing that RFFT might be a meta learning (learning to learn) ability. Through more advanced pre-training, models might have ``learned'' the rule following ability so that a few examples are enough to learn a new task.

\subsection{Failure Cases of RFFT and Scratchpad}\label{app:failure_RFFT_scratchpad}

\subsubsection{Scratchpad}
As discussed in Section~\ref{rule_result}, scratchpad struggles to learn the rationales behind each step without explicitly providing rules. We offer two failure examples in Table~\ref{tab:fail1_scratchpad} and ~\ref{tab:fail2_scratchpad}. Specifically, the model makes mistakes at the ``\# added 4+2+0=6'' step, indicating that it fails to locate the rightmost digit of the first number. Besides, the table~\ref{tab:fail2_scratchpad} shows that the model cannot correctly compute the carry C. These issues are likely due to the model cannot comprehend the principle behind the 'added' step. In contrast, when rules are clearly provided, the model is better equipped to understand the rationale and perform the rule-following process, thereby reducing the difficulty in the learning process.

\subsubsection{RFFT}
Although RFFT can teach transformers to do rule-based reasoning, the basic capabilities of LLMs might limit their strict generalization, leading to occasional errors during some basic operations. Refer to~Table~\ref{tab:fail_RFFT}. We observe that the model fails to output the correct digit popped out, resulting in its eventual incorrect answer.

\begin{table}[h]
\centering
\begin{tabularx}{\textwidth}{lXX}
\toprule
input: &
\multicolumn{2}{X}{
Follow the code step by step to answer the question:
\newline
[7, 8, 5, 3]+[8, 6, 8, 8, 1, 0]=
\newline
def sum\_digit\_by\_digit(num1, num2):
\newline
\hspace*{1em}\# Initialize the result list and carry
\newline
\hspace*{1em}result=[]
\newline
\hspace*{1em}carry=0
\newline
\hspace*{1em}...
} \\
\midrule
model output: & 
num1=[7, 8, 5, 3]
\newline
num2=[8, 6, 8, 8, 1, 0]
\newline
\textbf{1. Initialize Result and Carry}
\newline
result=[]
\newline
carry=0
\newline
\textbf{2. Loop Through Each Digit}
\newline
\verb|`|\verb|`|\verb|`|
\newline
\textit{while num1 or num2:}
\newline
\verb|`|\verb|`|\verb|`|
\newline
\textbf{2.1 check the stop criterion}
\newline
num1=[7,8,5,3]
\newline
num2=[8,6,8,8,1,0]
\newline
bool(num1)=True
\newline
bool(num2)=True
\newline
num1 or num2=True
\newline
enter the loop
\newline
\textbf{2.2 one iteration}
\newline
\verb|`|\verb|`|\verb|`|
\newline
\textit{digit1=num1.pop() if num1 else 0}
\newline
\verb|`|\verb|`|\verb|`|
\newline
num1=[7,8,5,3]
\newline
bool(num1)=True
\newline
num1.pop()
\newline
num1=[7,8,5]
\newline
digit1=3
\newline
...
\newline
result=[3]
\newline
\verb|`|\verb|`|\verb|`|
\newline
carry=total//10
\newline
\verb|`|\verb|`|\verb|`|
\newline
carry=3//10=0
\newline
...
\newline
\textbf{2.2 one iteration}
\newline
\verb|`|\verb|`|\verb|`|
\newline
\textit{digit1=num1.pop() if num1 else 0}
\newline
\verb|`|\verb|`|\verb|`|
\newline
num1=[7,8,5]
\newline
bool(num1)=True
\newline
num1.pop()
\newline
num1=[7,8]
\newline
digit1=5
&
\verb|`|\verb|`|\verb|`|
\newline
\textit{digit2=num2.pop() if num2 else 0}
\newline
\verb|`|\verb|`|\verb|`|
\newline
num2=[8,6,8,8,1]
\newline
bool(num2)=True
\newline
num2.pop()
\newline
num2=[8,6,8,8]
\newline
\textcolor{red}{digit2=8}
\newline
\verb|`|\verb|`|\verb|`|
\newline
\textit{total=digit1+digit2+carry}
\newline
\verb|`|\verb|`|\verb|`|
\newline
total=digit1+digit2+carry=5+8+0=13
\newline
\verb|`|\verb|`|\verb|`|
\newline
result.insert(0,total\%10)
\newline
\verb|`|\verb|`|\verb|`|
\newline
result=[3]
total\%10=13\%10=3
result=[3,3]
\newline
carry=13//10=1
\newline
...
\newline
\textbf{2.3 back to the start of the loop}
\newline
\verb|`|\verb|`|\verb|`|
\newline
\textit{while num1 or num2:}
\newline
\verb|`|\verb|`|\verb|`|
\newline
\textbf{2.1 check the stop criterion}
\newline
num1=[]
\newline
...
\newline
num1 or num2=False
\newline
end the loop
\newline
\textbf{3. Check Remaining Carry}
\newline
\verb|`|\verb|`|\verb|`|
\newline
\textit{if carry:}
\newline
\textit{\hspace*{1em}result.insert(0, carry)}
\newline
\verb|`|\verb|`|\verb|`|
\newline
result=[8,7,6,7,3,3]
\newline
carry=0
\newline
bool(carry)=False
\newline
pass
\newline
\textbf{4. Return Result}
\newline
\verb|`|\verb|`|\verb|`|
\newline
\textit{return result}
\newline
\verb|`|\verb|`|\verb|`|
\newline
result=[8,7,6,7,3,3]
\\\bottomrule
\end{tabularx}
\caption{The generation of fine-tuned Llama-2-7B with RFFT.}\label{tab:fail_RFFT}
\end{table}

\subsection{Rule-following with Code Representations and Natural Language Representations}
\label{app:code_nl}
We offer two types of rule-following input-output sequences including code representations and natural language representations to show that rules can be of various formats. We provide examples of full input-output sequences for reference in Appendix~\ref{full_IO}. The results are shown in Figure~\ref{fig:nl_code}.

Besides, we will discuss the difference of our work from previous work~\citet{nye2021scratchpad} which teaches LLMs to execute code (which they call scratchpad tracing) as follows. Our RFFT aims at teaching LLMs to follow explicitly provided rules to reason rather than to execute programs like an interpreter. We show in Figure~\ref{fig:nl_code} that rule can be of various forms including programs and natural language. 
Besides, we provide detailed natural language instructions in the rule-following input-output sequences with code representations. We expect the instructions may help LLMs to recall knowledge learned in the pre-training corpus.

\begin{figure}[H]
    \centering
    \includegraphics[width=.4\textwidth]{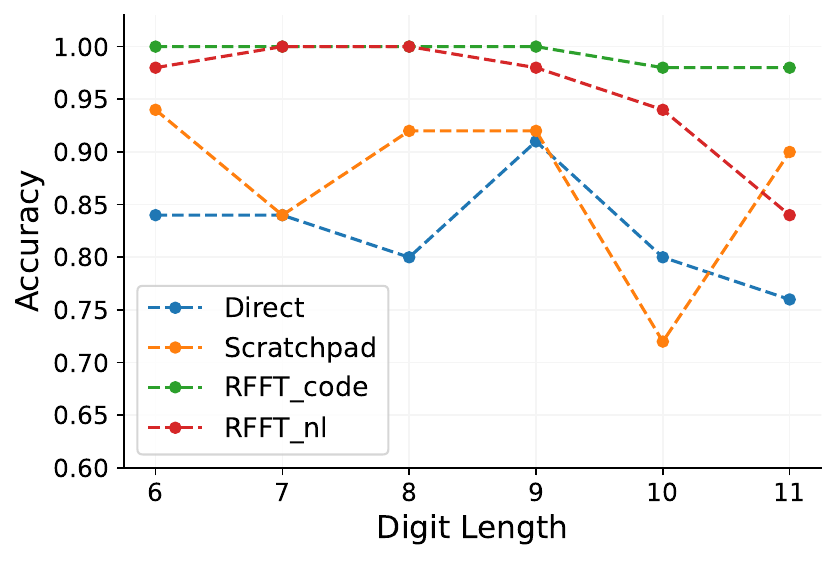}
    \caption{Test accuracy of GPT-3.5-turbo fine-tuned on 1-5 digit addition. We train the model in 4 different settings. The full input-output pair can be found in Appendix~\ref{full_IO}.}
    \label{fig:nl_code}
\end{figure}

\subsection{Comparing RFFT with Scratchpad Tracing}

We propose the technique of RFFT in~\S\ref{sec:experiment_rule}, which instructs LLMs to follow rules of various forms. In~\S\ref{sec:experiment_rule}, we use rules represented by programs; in Appendix~\ref{app:code_nl}, we provide another option of rules represented by natural language. A related work~\citet{nye2021scratchpad} introduces a method of fine-tuning LLMs to predict the program execution trace line by line, called ``scratchpad tracing''. We here state the difference between scratchpad tracing and our RFFT with programs as rules: 1) we provide the detailed execution process of each line of the code instead of directly giving the value of variables after line-by-line execution, through which we decompose each step of execution in a more fine-grained way; 2) we provide natural language instructions or the rationales behind each step to help the LLMs to understand the execution steps, for example ``1. Initialize Result and Carry'', ``2. Loop Through Each Digit'', etc. In summary, RFFT with programs teaches LLMs to execute code in a more human-readable way, simulating how human read, understand, and execute the code in their mind, while scratchpad tracing directly predicts program traces using raw machine formats.

{To further demonstrate the effectiveness of our technique, we use scratchpad tracing to fine-tune GPT-3.5-turbo-1106 on the addition task. We still maintain the same data size and training parameters as RFFT, i.e., 100 training samples with 4 epochs and batch size of 4. The full input-output pair is provided in Appendix~\ref{full_IO}. Considering the expensive cost with OpenAI API, we generate 100 test samples for each digit length and perform three independent experiments and report average accuracy and standard deviation. We show the results of RFFT and scratchpad tracing in Figure~\ref{fig:gpt3_ft_trace}. RFFT significantly outperforms scratchpad tracing. The results show that the detailed execution process of each line of code and natural language instructions enhance the model's rule learning ability and improve length generalization.

\begin{figure}[]
    \centering
    \includegraphics[width=.4\textwidth]{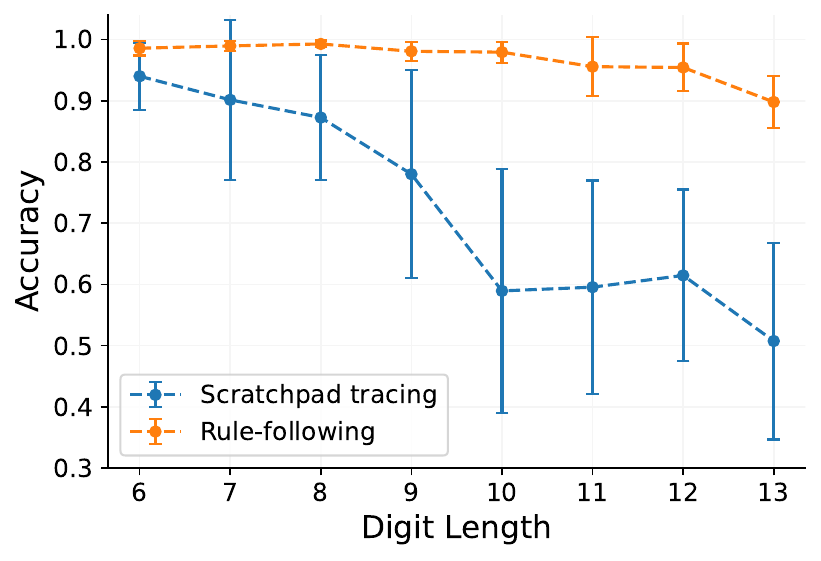}
    \caption{Results of RFFT and scratchpad tracing on addition.}
    \label{fig:gpt3_ft_trace}
\end{figure}

\subsection{RFFT on Other Tasks}
\label{sec:last_letter}
Besides addition shown in~\S\ref{sec:experiment_rule}, we conduct experiments of RFFT on two additional tasks including base-9 addition and last letter concatenation. Last letter concatenation is introduced in~\citet{few-shot-cot}. In the task, the model is asked to concatenate the last letters of words. We choose the words from top one-thousand last names from \url{https://namecensus.com/}.

We fine-tune GPT-3.5-turbo-1106 with a training set of 100 samples of 1-5 digit base addition or of concatenating the last letter of 1-5 words respectively for two tasks. We train the models with OpenAI API with 4 epochs, 4 batch sizes like in Appendix~\ref{rule_detail}. Then, we test the model on test samples of 6-15 length. We list the full prompt for last letter concatenation in Appendix~\ref{prompt_last_letter}, as the prompt for base addition is basically the same as that for addition.

The results are shown in Figure~\ref{fig:rfft_on_other_task}. RFFT outperforms the method of direct answer and scratchpad significantly, showing that RFFT enhances the model ability of following given rules to solve problems.

\begin{figure}[t]
  \centering
  \subfigure{
    \label{fig:gpt3_ft_base_addition} 
    \includegraphics[width=.4\textwidth]{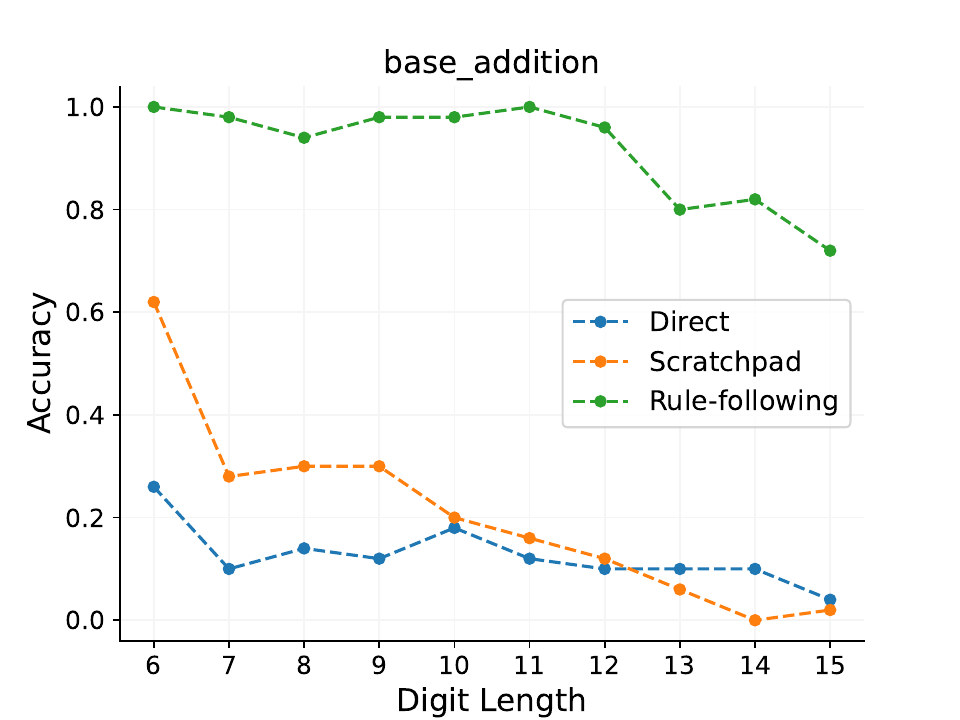}}
  \subfigure{    \includegraphics[width=.4\textwidth]{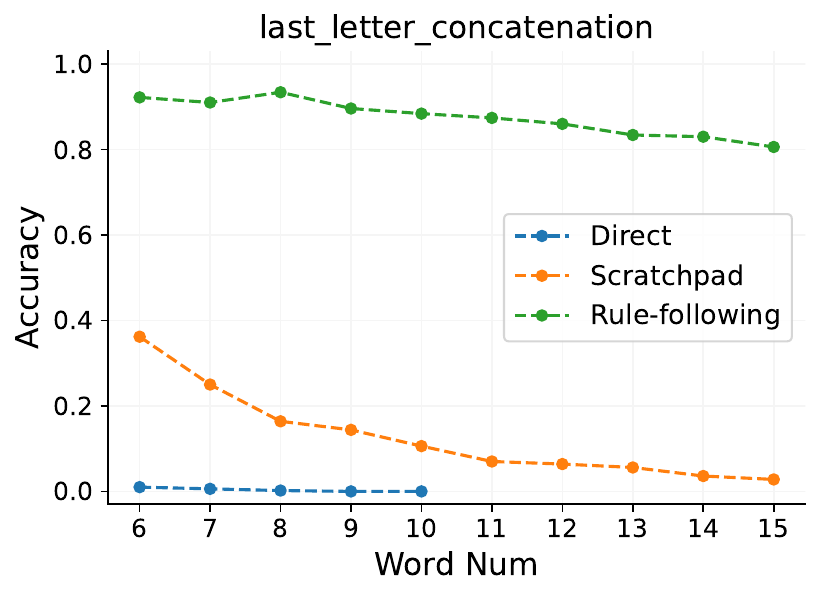}
    \label{fig:gpt3_ft_last_letter}}
\caption{Results of direct, scratchpad and RFFT on \textbf{base addition} (\textbf{left}) and \textbf{last letter concatenation} (\textbf{right}).}
\label{fig:rfft_on_other_task}
\end{figure}

\section{Scratchpad vs Direct Answer on Addition}
\label{app: scratch_vs_direct}
We increase the number of training samples for scratchpad to 5,000 on the task of addition with GPT-3.5-turbo. In detail, we average the results over 3 models fine-tuned for 1 epoch respectively. We use a test set of 100 samples. As is shown in Figure~\ref{fig:gpt3_ft_more_sample}, scratchpad with 5,000 training samples outperforms direct answer with 100 samples but is still worse than RFFT with 100 samples.

\begin{figure}[H]
    \centering
    \includegraphics[width=.4\textwidth]{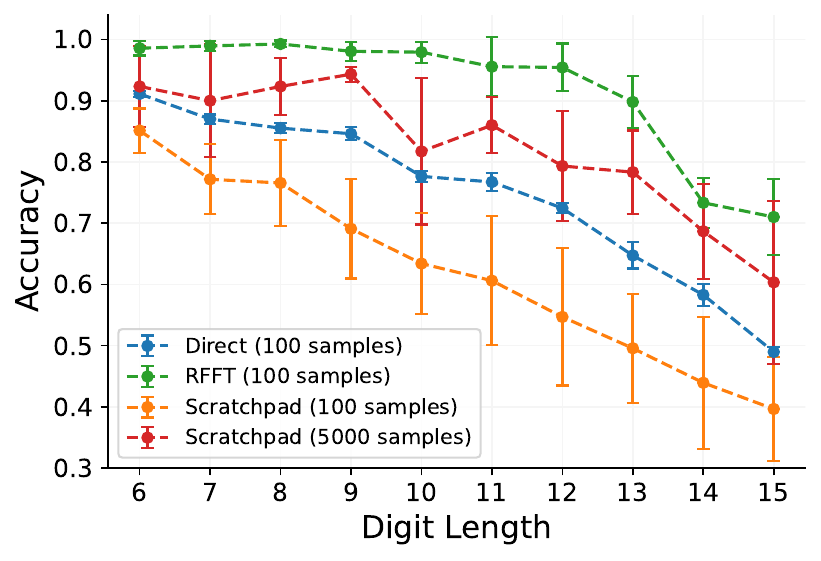}
    \caption{Results of direct answer with 100 training samples, scratchpad with 100 and 5,000 training samples respectively.}
    \label{fig:gpt3_ft_more_sample}
\end{figure}

\section{Experiments of Grokking}
\label{grokking}
~\citet{nanda2023progress, zhong2023pizza} have claimed that transformers can learn systematic rules to solve modular addition. They show through experiments that transformers are embedding numbers as angles (points on the unit circles) and complete modular additions by operating on trigonometric functions of the angles. We perform the Leave-Square-Out method in the same settings as in~\citet{zhong2023pizza}. The only change is the training-test data split. Specifically, the task is $a+b \mod 59=$, $a\in[0, 58],b\in[0,58]$. We leave a square test set of side length $l_k=16$ (8\% of the whole set) out and train 5 transformers with the same setting, while~\citet{zhong2023pizza} split the dataset randomly with training set accounting for 80\% of the whole dataset. We show that holes still appear, indicating that even such ability to learn the systematic algorithms and apply them to unseen samples rely severely on seeing similar cases. We describe the experiment in detail as follows.

The task is $a+b\mod 59 =$, $a\in[0, 58], b\in[0, 58]$. We leave a test square of length $l_k=16$ and center $(29, 29)$ out. Our training set accounts for about 92\% of the whole dataset while the training set accounts for 80\% in~\citet{zhong2023pizza}. The model can achieve 100\% accuracy when the dataset is randomly split with training set accounting for 80\%. This shows that the size of our training set is entirely sufficient for the model to solve the problem. Besides, we use the same hyper-parameters and model settings as given in the code of~\citet{zhong2023pizza}. We list them in Table~\ref{tab:grokking_setting}.

The results are shown in Figure~\ref{fig:grokking}. There are holes in the test square, indicating that the model can not perform well in the test square. This shows that even in the settings where grokking happens, the ability to learn the systematic rules and apply it to test samples may still rely on seeing similar cases.

\begin{figure}[H]
    \centering
    \includegraphics[width=\textwidth]{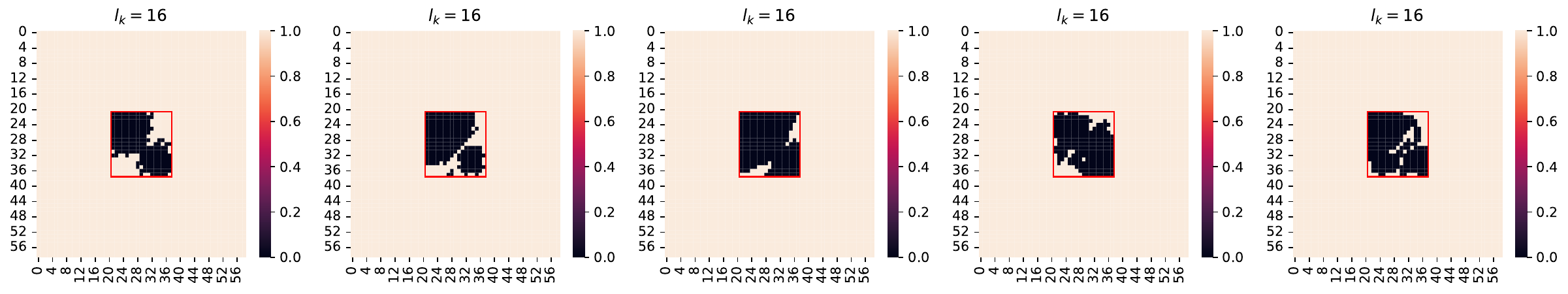}
    \caption{Accuracy distribution of the whole dataset when we leave a test square out. The test square is of length $l_k=16$ and center $(29, 29)$. We repeat the experiment for 5 times with the same settings.}
    \label{fig:grokking}
\end{figure}

\begin{table}[H]
    \centering
    \begin{tabular}{c|c}
    \toprule
     epoch  & 20000 \\
     batch size & 3481 \\
     learning rate & 0.001 \\
     weight decay & 2.0 \\
     model dimension & 128 \\
     n\_head & 4 \\
     n\_layer & 1 \\
    \bottomrule
    \end{tabular}
    \caption{Hyperparameters and model details of the experiments of grokking.}
    \label{tab:grokking_setting}
\end{table}



\section{Full Input-Output Sequences}
\label{full_IO}
In this section, we list the full input-output pairs used in~\S\ref{sec:experiment_rule} and in Appendix~\ref{sec:last_letter}.

\subsection{Addition}
We list the input-output sequences of addition as follows.
\paragraph{Direct answer}
We list an example of the input-output sequences for direct answer in Table~\ref{prompt_direct_answer}.
\paragraph{Scratchpad}
We list an example of the input-output sequences for scratchpad in Table~\ref{prompt_scratchpad}.

\paragraph{Scratchpad tracing}
We list an example of the input-output sequences for scratchpad tracing Table~\ref{prompt_scratchpad_tracing}.

\paragraph{Rule-following}
We list an example of the input for rule-following in Table~\ref{prompt_rule-following_input} and an example of the output in Table~\ref{prompt_rule-following_output}.

\paragraph{Rule-following with natural language representations}
We list an example of the input for rule-following with NL representations in Table~\ref{prompt_rule-following_input_nl} and an example of the output in Table~\ref{prompt_rule-following_output_nl}.

\subsection{Last letter concatenation}
\label{prompt_last_letter}
We list the input-output sequences of last letter concatenation as follows.
\paragraph{Direct answer}
We list an example of the input-output sequences for direct answer in Table~\ref{prompt_direct_answer_last_letter}.

\paragraph{Scratchpad}
We list an example of the input-output sequences for scratchpad in Table~\ref{prompt_scratchpad_last_letter}.

\paragraph{Rule-following}
We list an example of the input for rule-following in Table~\ref{prompt_rule-following_input_last_letter} and an example of the output in Table~\ref{prompt_rule-following_output_last_letter}.

\begin{table}[h]
\centering
\begin{tabular}{ll}
\toprule
input: & 7+6=  \\
\midrule
output: & 13\\
\bottomrule
\end{tabular}
\caption{Input-output sequence for direct answer on addition.}
\label{prompt_direct_answer}
\end{table}

\begin{table}[h]
\centering
\begin{tabularx}{.3\textwidth}{lX}
\toprule
input: & 7+6=  \\
\midrule
output: & 
7+6,,C:0
\newline
\# added 7+6+0=3\newline
+,3,C:1\newline
\# added 0+0+1=1\newline
+,13,C:0\newline
13\\
\bottomrule
\end{tabularx}
\caption{Input-output sequence for scratchpad on addition.}
\label{prompt_scratchpad}
\end{table}

\begin{table}[h]
\centering
\begin{tabularx}{\textwidth}{lX}
\toprule
input: & 
def sum\_digit\_by\_digit(num1, num2):
\newline
\hspace*{1em}result=[]
\newline
\hspace*{1em}carry=0
\newline
\hspace*{1em}while num1 or num2:
\newline
\hspace*{1em}\hspace*{1em}digit1=num1.pop() if num1 else 0
\newline
\hspace*{1em}\hspace*{1em}digit2=num2.pop() if num2 else 0
\newline
\hspace*{1em}\hspace*{1em}total=digit1+digit2+carry
\newline
\hspace*{1em}\hspace*{1em}result.insert(0,total\%10)
\newline
\hspace*{1em}\hspace*{1em}carry=total//10
\newline
\hspace*{1em}if carry:
\newline
\hspace*{1em}\hspace*{1em}result.insert(0, carry)
\newline
\hspace*{1em}return result
\newline
\newline
output = sum\_digit\_by\_digit([7, 3], [4, 8]) \\
\midrule
output: & 
state: \{\}
\newline
line: def sum\_digit\_by\_digit(num1, num2):
\newline
state: \{`sum\_digit\_by\_digit': `\textless callable\_object sum\_digit\_by\_digit\textgreater'\}
\newline
line:     result=[]
\newline
state: \{`num1': [7, 3], `num2': [4, 8], `result': []\}
\newline
line:     carry=0
\newline
state: \{`num1': [7, 3], `num2': [4, 8], `result': [], `carry': 0\}
\newline
line:     while num1 or num2:
\newline
state: \{`num1': [7, 3], `num2': [4, 8], `result': [], `carry': 0\}
\newline
line:         digit1=num1.pop() if num1 else 0
\newline
state: \{`num1': [7], `num2': [4, 8], `result': [], `carry': 0, `digit1': 3\}
\newline
line:         digit2=num2.pop() if num2 else 0
\newline
state: \{`num1': [7], `num2': [4], `result': [], `carry': 0, `digit1': 3, 'digit2': 8\}
\newline
line:         total=digit1+digit2+carry
\newline
state: \{`num1': [7], `num2': [4], `result': [], `carry': 0, `digit1': 3, `digit2': 8, `total': 11\}
\newline
line:         result.insert(0,total\%10)
\newline
state: \{`num1': [7], `num2': [4], `result': [1], `carry': 0, `digit1': 3, `digit2': 8, `total': 11\}
\newline
line:         carry=total//10
\newline
state: \{`num1': [7], `num2': [4], `result': [1], `carry': 1, `digit1': 3, `digit2': 8, `total': 11\}
\newline
line:     while num1 or num2:
\newline
state: \{`num1': [7], `num2': [4], `result': [1], `carry': 1, `digit1': 3, `digit2': 8, `total': 11\}
\newline
line:         digit1=num1.pop() if num1 else 0
\newline
state: \{`num1': [], `num2': [4], `result': [1], `carry': 1, `digit1': 7, `digit2': 8, `total': 11\}
\newline
line:         digit2=num2.pop() if num2 else 0
\newline
state: \{`num1': [], `num2': [], `result': [1], `carry': 1, `digit1': 7, `digit2': 4, `total': 11\}
\newline
line:         total=digit1+digit2+carry
\newline
state: \{`num1': [], `num2': [], `result': [1], `carry': 1, `digit1': 7, `digit2': 4, `total': 12\}
\newline
line:         result.insert(0,total\%10)
\newline
state: \{`num1': [], `num2': [], `result': [2, 1], `carry': 1, `digit1': 7, `digit2': 4, `total': 12\}
\newline
line:         carry=total//10
\newline
state: \{`num1': [], `num2': [], `result': [2, 1], `carry': 1, `digit1': 7, `digit2': 4, `total': 12\}
\newline
line:     if carry:
\newline
state: \{`num1': [], `num2': [], `result': [2, 1], `carry': 1, `digit1': 7, `digit2': 4, `total': 12\}
\newline
line:         result.insert(0, carry)
\newline
state: \{`num1': [], `num2': [], `result': [1, 2, 1], `carry': 1, `digit1': 7, `digit2': 4, `total': 12\}
\newline
line:     return result
\newline
state: \{`sum\_digit\_by\_digit': `\textless callable\_object sum\_digit\_by\_digit\textgreater', `output': [1, 2, 1]\}
\\
\bottomrule
\end{tabularx}
\caption{Input-output sequence for scratchpad tracing on addition.}
\label{prompt_scratchpad_tracing}
\end{table}

\begin{table}[h]
\centering
\begin{tabularx}{\textwidth}{lX}
\toprule
input: &
Follow the code step by step to answer the question:
\newline
[7]+[6]=
\newline
def sum\_digit\_by\_digit(num1, num2):
\newline
\hspace*{1em}\# Initialize the result list and carry
\newline
\hspace*{1em}result=[]
\newline
\hspace*{1em}carry=0
\newline
\hspace*{1em}\# Loop through each digit
\newline
\hspace*{1em}while num1 or num2:
\newline
\hspace*{1em}\hspace*{1em}\# Get the current digits, defaulting to 0 if one number is shorter
\newline
\hspace*{1em}\hspace*{1em}digit1=num1.pop() if num1 else 0
\newline
\hspace*{1em}\hspace*{1em}digit2=num2.pop() if num2 else 0
\newline
\hspace*{1em}\hspace*{1em}\# Calculate the sum of the current digits and the carry
\newline
\hspace*{1em}\hspace*{1em}total=digit1+digit2+carry
\newline
\hspace*{1em}\hspace*{1em}\# Insert the last digit of total to the beginning of the result and update carry
\newline
\hspace*{1em}\hspace*{1em}result.insert(0,total\%10)
\newline
\hspace*{1em}\hspace*{1em}carry=total//10
\newline
\hspace*{1em}\# If there's a remaining carry, insert it to the beginning of the result
\newline
\hspace*{1em}if carry:
\newline
\hspace*{1em}\hspace*{1em}result.insert(0, carry)
\newline
\hspace*{1em}\# Return the result
\newline
\hspace*{1em}return result\\
\bottomrule
\end{tabularx}
\caption{Input for rule-following on addition.}
\label{prompt_rule-following_input}
\end{table}

\begin{table}[h]
\centering
\begin{tabularx}{\textwidth}{lXX}
\toprule
output: & 
num1=[7]
\newline
num2=[6]
\newline
\textbf{1. Initialize Result and Carry}
\newline
result=[]
\newline
carry=0
\newline
\textbf{2. Loop Through Each Digit}
\newline
\verb|`|\verb|`|\verb|`|
\newline
\textit{while num1 or num2:}
\newline
\verb|`|\verb|`|\verb|`|
\newline
\textbf{2.1 check the stop criterion}
\newline
num1=[7]
\newline
num2=[6]
\newline
bool(num1)=True
\newline
bool(num2)=True
\newline
num1 or num2=True
\newline
enter the loop
\newline
\textbf{2.2 one iteration}
\newline
\verb|`|\verb|`|\verb|`|
\newline
\textit{digit1=num1.pop() if num1 else 0}
\newline
\verb|`|\verb|`|\verb|`|
\newline
num1=[7]
\newline
bool(num1)=True
\newline
num1.pop()
\newline
num1=[]
\newline
digit1=7
\newline
\verb|`|\verb|`|\verb|`|
\newline
\textit{digit2=num2.pop() if num2 else 0}
\newline
\verb|`|\verb|`|\verb|`|
\newline
num2=[6]
\newline
bool(num2)=True
\newline
num2.pop()
\newline
num2=[]
\newline
digit2=6
\newline
\verb|`|\verb|`|\verb|`|
\newline
\textit{total=digit1+digit2+carry}
\newline
\verb|`|\verb|`|\verb|`|
\newline
total=digit1+digit2+carry=7+6+0=13
&
\verb|`|\verb|`|\verb|`|
\newline
\textit{result.insert(0,total\%10)}
\newline
\verb|`|\verb|`|\verb|`|
\newline
result=[]
\newline
total\%10=13\%10=3
\newline
result=[3]
\newline
\verb|`|\verb|`|\verb|`|
\newline
\textit{carry=total//10}
\newline
\verb|`|\verb|`|\verb|`|
\newline
carry=13//10=1
\newline
\textbf{2.3 back to the start of the loop}
\newline
\verb|`|\verb|`|\verb|`|
\newline
\textit{while num1 or num2:}
\newline
\verb|`|\verb|`|\verb|`|
\newline
\textbf{2.1 check the stop criterion}
\newline
num1=[]
\newline
num2=[]
\newline
bool(num1)=False
\newline
bool(num2)=False
\newline
num1 or num2=False
\newline
end the loop
\newline
\textbf{3. Check Remaining Carry}
\newline
\verb|`|\verb|`|\verb|`|
\newline
\textit{if carry:}
\newline
\textit{\hspace*{1em}result.insert(0, carry)}
\newline
\verb|`|\verb|`|\verb|`|
\newline
result=[3]
\newline
carry=1
\newline
bool(carry)=True
\newline
result=[1,3]
\newline
\textbf{4. Return Result}
\newline
\verb|`|\verb|`|\verb|`|
\newline
\textit{return result}
\newline
\verb|`|\verb|`|\verb|`|
\newline
result=[1,3] \\\bottomrule
\end{tabularx}
\caption{Output for rule-following on addition.}
\label{prompt_rule-following_output}
\end{table}

\begin{table}[h]
\centering
\begin{tabularx}{.55\textwidth}{lX}
\toprule
input: & 
Concatenate the last letter of the following names:
\newline
King, Kaur  \\
\midrule
output: & gr\\
\bottomrule
\end{tabularx}
\caption{Input-output sequence for direct answer on last letter concatenation.}
\label{prompt_direct_answer_last_letter}
\end{table}

\begin{table}[h]
\centering
\begin{tabularx}{.55\textwidth}{lX}
\toprule
input: & 
Concatenate the last letter of the following names:
\newline
King, Kaur  \\
\midrule
output: & 
\# Concatenating the last letter of King
\newline
g
\newline
\# Concatenating the last letter of Kaur
\newline
gr\\
\bottomrule
\end{tabularx}
\caption{Input-output sequence for scratchpad on last letter concatenation.}
\label{prompt_scratchpad_last_letter}
\end{table}


\begin{table}[h]
\centering
\begin{tabularx}{.8\textwidth}{lX}
\toprule
input: &
Follow the code step by step to concatenate the last letter of the following names:
\newline
King, Kaur
\newline
def last\_letter\_concat(names):
\newline
\hspace*{1em}\# Initialize Result
\newline
\hspace*{1em}result = ""
\newline
\hspace*{1em}\# Main Loop
\newline
\hspace*{1em}for name in names:
\newline
\hspace*{1em}\hspace*{1em}result += name[-1]
\newline
\hspace*{1em}return result\\
\bottomrule
\end{tabularx}
\caption{Input for rule-following on last letter concatenation.}
\label{prompt_rule-following_input_last_letter}
\end{table}

\begin{table}[h]
\centering
\begin{tabularx}{\textwidth}{lXX}
\toprule
output: 
& 
names = ['King','Kaur']
\newline
\textbf{1. Initialze result}
\newline
result = ""
\newline
\textbf{2. Main Loop}
\newline
\textbf{2.1 one iteration}
\newline
\verb|`|\verb|`|\verb|`|
\newline
for name in names:
\newline
\verb|`|\verb|`|\verb|`|
\newline
name = "King"
\newline
\verb|`|\verb|`|\verb|`|
\newline
result += name[-1]
\newline
\verb|`|\verb|`|\verb|`|
\newline
result = ""
\newline
name[-1] = "g"
\newline
result += "g"
\newline
result = "g"
&
\textbf{2.1 one iteration}
\newline
\verb|`|\verb|`|\verb|`|
\newline
for name in names:
\newline
\verb|`|\verb|`|\verb|`|
\newline
name = "Kaur"
\newline
\verb|`|\verb|`|\verb|`|
\newline
result += name[-1]
\newline
\verb|`|\verb|`|\verb|`|
\newline
result = "g"
\newline
name[-1] = "r"
\newline
result += "r"
\newline
result = "gr"
\newline
\textbf{3. Return Result}
\newline
\verb|`|\verb|`|\verb|`|
\newline
resturn result
\newline
\verb|`|\verb|`|\verb|`|
\newline
result = "gr"
\\\bottomrule
\end{tabularx}
\caption{Output for rule-following on last letter concatenation.}
\label{prompt_rule-following_output_last_letter}
\end{table}

\begin{table}[h]
\centering
\begin{tabularx}{\textwidth}{lX}
\toprule
input: &
Follow the rules step by step to answer the question:
\verb|`|6\verb|`|+\verb|`|7\verb|`|=
\newline
Add two numbers in order from the lowest digit to the highest digit. The operation rules are as follows:
\newline
1. In the initial state, the carry from the previous digit is 0 and the result is \textless empty\textgreater.
\newline
2. Begin the loop through each digit:
\newline
\hspace*{1em}2.1. Determine whether the two numbers are \textless empty\textgreater. If both are \textless empty\textgreater, the loop is over. If one of them is not \textless empty\textgreater, go to the iteration.
\newline
\hspace*{1em}2.2. One iteration:
\newline
\hspace*{1em}\hspace*{1em}2.2.1. Find the rightmost digit of each number and remove the digit from the number. If one of them is \textless empty\textgreater, the digit is 0.
\newline
\hspace*{1em}\hspace*{1em}2.2.2. Calculate the sum of the current digits and the carry.
\newline
\hspace*{1em}\hspace*{1em}2.2.3. Insert the last digit of the summation to the beginning of the result and update carry. If the sum is larger than 9, carry is 1. Otherwise, carry is 0.
\newline
\hspace*{1em}2.3. Back to the start of the loop.
\newline
3. When the loop through each digit is over, if there's a remaining carry, insert it to the beginning of the result.
\newline
4. Return the result.
\\\bottomrule
\end{tabularx}
\caption{Input for rule-following-NL.}
\label{prompt_rule-following_input_nl}
\end{table}

\begin{table}[h]
\centering

\renewcommand\arraystretch{0.5}
\begin{tabularx}{\textwidth}{lXX}
\toprule
output: & 
num1=\verb|`|6\verb|`|;
num2=\verb|`|7\verb|`|
\newline
\verb|`|\verb|`|\verb|`|
\newline
1. In the initial state, the carry from the previous digit is 0 and the result is \textless empty\textgreater.
\newline
\verb|`|\verb|`|\verb|`|
\newline
Initialize the carry as 0 and the result is \verb|`|\textless empty\textgreater\verb|`|.
\newline 
\verb|`|\verb|`|\verb|`|
\newline
2. Begin the loop through each digit:
\newline
2.1. Determine whether the two numbers are \textless empty\textgreater. If both are \textless empty\textgreater, the loop is over. If one of them is not \textless empty\textgreater, go to the iteration.
\newline 
\verb|`|\verb|`|\verb|`|
\newline
num1 \verb|`|6\verb|`| is not \textless empty\textgreater.
num2 \verb|`|7\verb|`| is not \textless empty\textgreater.
\newline
Because both are not \textless empty\textgreater, go to the iteration.
\newline 
\verb|`|\verb|`|\verb|`|
\newline
2.2. One iteration:
\newline
2.2.1. Find the rightmost digit of each number and remove the digit from the number. If one of them is \textless empty\textgreater, the digit is 0.
\newline 
\verb|`|\verb|`|\verb|`|
\newline
The rightmost digit of num1 \verb|`|6\verb|`| is \verb|`|6\verb|`|. The rightmost digit of num2 \verb|`|7\verb|`| is \verb|`|7\verb|`|.
Then the remaining num1 \verb|`|6\verb|`| is \verb|`|\textless empty\textgreater\verb|`| and the remaining num2 \verb|`|7\verb|`| is \verb|`|\textless empty\textgreater\verb|`|.
\newline
\verb|`|\verb|`|\verb|`|
\newline
2.2.2. Calculate the sum of the current digits and the carry.
\newline 
\verb|`|\verb|`|\verb|`|
\newline
The sum of \verb|`|6\verb|`|, \verb|`|7\verb|`| and carry \verb|`|0\verb|`| is 6 + 7 + 0 = 13.
\newline
\verb|`|\verb|`|\verb|`|
\newline
2.2.3. Insert the last digit of the summation to the beginning of the result and update carry.
\newline 
\verb|`|\verb|`|\verb|`|
\newline
The last digit of the summation is 13 \% 10 = 3. The previous result is \verb|`|\textless empty\textgreater\verb|`|, now the result is: \verb|`|3\verb|`|.
\newline
The carry is 13 // 10 = 1.
\newline
\verb|`|\verb|`|\verb|`|
\newline
2.3. Back to the start of the loop.
\verb|`|\verb|`|\verb|`|
\newline
\verb|`|\verb|`|\verb|`|
\newline
2.1. Determine whether the two numbers are \textless empty\textgreater. If both are \textless empty\textgreater, the loop is over. If one of them is not \textless empty\textgreater, go to the iteration.\newline
\verb|`|\verb|`|\verb|`|
\newline
num1 is \verb|`|\textless empty\textgreater\verb|`|.
num2 is \verb|`|\textless empty\textgreater\verb|`|.
\newline
Because both are \textless empty\textgreater, the loop is over.
\newline
\verb|`|\verb|`|\verb|`|
\newline
3. When the loop through each digit is over, if there's a remaining carry, insert it to the beginning of the result.
\verb|`|\verb|`|\verb|`|
\newline
The carry is 1, so add an 1 to the beginning of the result \verb|`|3\verb|`|. So the final result is \verb|`|1,3\verb|`|.
\newline
\verb|`|\verb|`|\verb|`|
\newline
4. Return the result.
\verb|`|\verb|`|\verb|`|
\newline
The final result is \verb|`|1,3\verb|`|.
\\\bottomrule
\end{tabularx}
\caption{Output for rule-following-nl.}
\label{prompt_rule-following_output_nl}
\end{table}

\end{document}